\documentclass{article}

\usepackage{PRIMEarxiv}

\usepackage[utf8]{inputenc} 
\usepackage[T1]{fontenc}    
\usepackage{hyperref}       
\usepackage{url}            
\usepackage{booktabs}       
\usepackage{amsfonts}       
\usepackage{nicefrac}       
\usepackage{microtype}      
\usepackage{lipsum}
\usepackage{fancyhdr}       
\usepackage{graphicx}       
\usepackage{subfigure}
\usepackage{amsmath}
\graphicspath{{media/}}     

\title{Graph Structure Prompt Learning: A Novel Methodology to Improve Performance of Graph Neural Networks
}

\author{
  Zhenhua Huang, Kunhao Li, Shaojie Wang, Zhaohong Jia \\
  School of Internet \\
  Anhui University \\
  Hefei\\
  \texttt{\{zhhuangscut, kunhomlihf, wsj.ahu\}@gmail.com, zhjia@mail.ustc.edu.cn} \\
   \And
  Wentao Zhu \\
  Amazon Research \\
  Seattle\\
  \texttt{wentaozhu91@gmail.com} \\
   \And
  Sharad Mehrotra \\
  University of California Irvine\\
  Irvine \\
  \texttt{sharad@ics.edu}\\
}

\begin{document}
\maketitle

\begin{abstract}
Graph neural networks (GNNs) are widely applied in graph data modeling. However, existing GNNs are often trained in a task-driven manner that fails to fully capture the intrinsic nature of the graph structure, resulting in sub-optimal node and graph representations. To address this limitation, we propose a novel \textbf{G}raph structure \textbf{P}rompt \textbf{L}earning method (GPL) to enhance the training of GNNs, which is inspired by prompt mechanisms in natural language processing. GPL employs task-independent graph structure losses to encourage GNNs to learn intrinsic graph characteristics while simultaneously solving downstream tasks, producing higher-quality node and graph representations. In extensive experiments on eleven real-world datasets, after being trained by GPL, GNNs significantly outperform their original performance on node classification, graph classification, and edge prediction tasks (up to 10.28\%, 16.5\%, and 24.15\%, respectively). By allowing GNNs to capture the inherent structural prompts of graphs in GPL, they can alleviate the issue of over-smooth and achieve new state-of-the-art performances, which introduces a novel and effective direction for GNN research with potential applications in various domains.
\end{abstract}


\section{Introduction}
Many real-world datasets are presented as networks or graphs, and various types of graph neural networks (GNNs) \cite{gnnreview1} have been developed to address the inherent challenges presented by these datasets. GNNs have gained significant attention in various fields of data mining applications, such as knowledge representation \cite{knowledge2021, 2022exgnn}, text classification \cite{textc2019, 2020heterogeneous4doc}, traffic prediction \cite{traffic2021,trafficprediction2022}, molecular classification \cite{molecular2018, 2023gambgnn}, recommendation systems \cite{2023ipm_gnn4recommand}, sentiment analysis \cite{2022sent_analy,sentiment2023},~\textit{etc}. Classic and advanced graph neural networks, including ChebNet \cite{gnn2016}, graph convolution networks (GCN) \cite{gcn2017}, GraphSage \cite{sage2017}, graph attention networks (GAT) \cite{gat2018}, LightGCN \cite{lgconv2020}, UniMP \cite{transformconv2021}, ARMA \cite{ARMA2021}, Fused GAT \cite{2022fusedgat}, ASDGN \cite{2022anti-symmetric} ~\textit{etc.}, were initially designed for node classification tasks. To gather node features for graph classification tasks, various readout functions and pooling mechanisms have been proposed, such as GIN \cite{gin2018}, SortPool \cite{sortpool2018}, DiffPool \cite{diffpool2019}, TopKPool \cite{topkpool2019}, SAGPool \cite{sagpool2019}, EdgePool \cite{edgepool2019}, ASAPool \cite{asapool2020}, and MEWISPool \cite{MEWISPool2021}, GPS \cite{GPS2022}, ~\textit{etc}. Compared with the thriving models and applications, training methods for improving the performance of the classic graph neural networks are less discussed. Several works use graph pre-train technologies or heterogeneous attributes to enhance the performance of certain GNNs \cite{hu2019strategies,gcc2020,gptgnn2020, L2P2021,sugar2021,metagnn}, but these methods heavily rely on extra graph datasets or skillfully designed graph patterns, and their improved performances and generalization are also limited in certain cases.
\begin{figure*}[h]
  \centering
    \includegraphics[width=\linewidth]{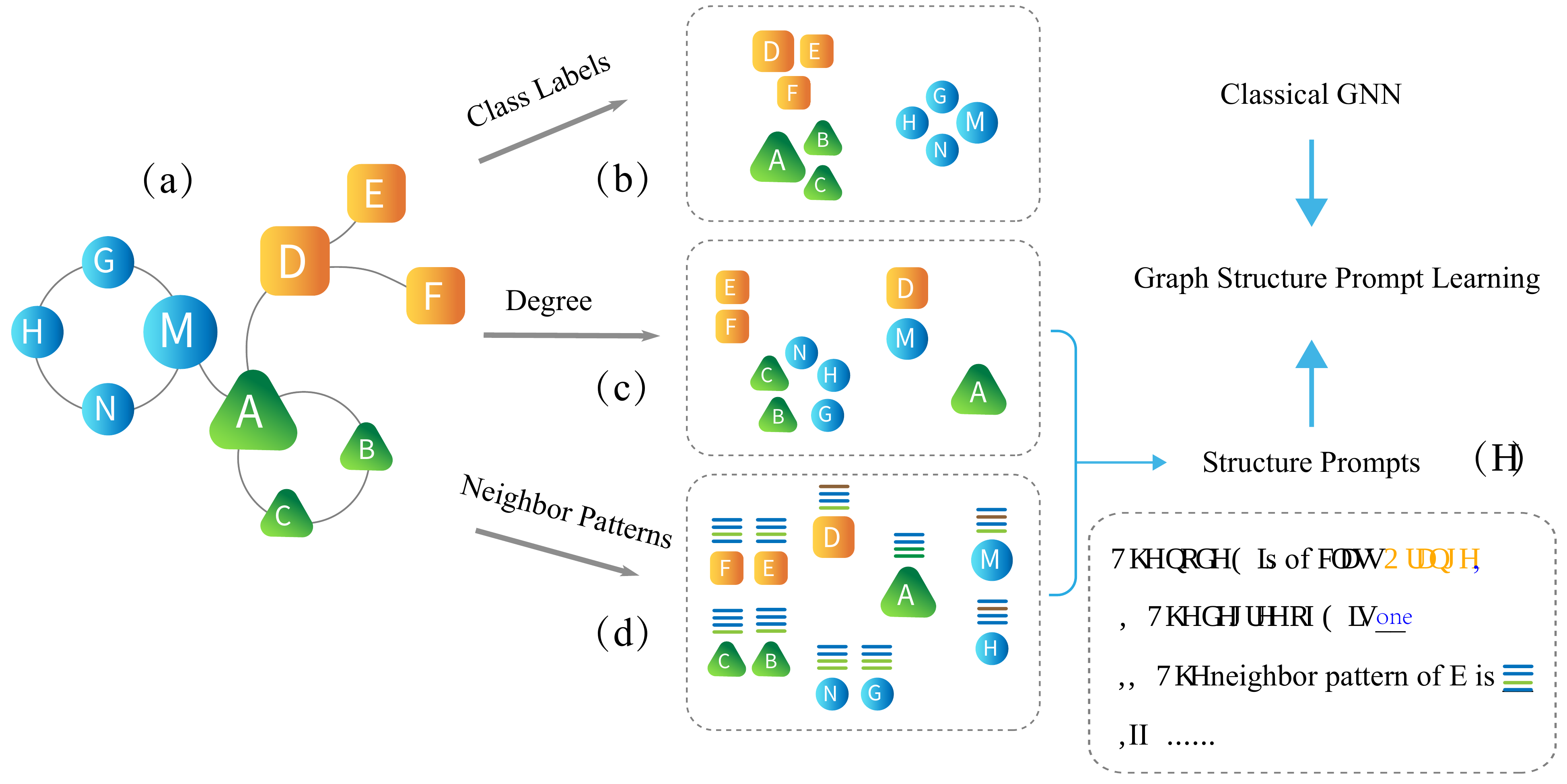}
  \caption{A running example of graph structure prompt learning. The color and size represent nodes' labels and degrees, respectively. (a) The example graph. (b) Nodes clustered by class labels. (c) Nodes clustered by degrees. (d) Nodes clustered by neighboring patterns (degree distributions of neighboring nodes in four-dimension). (e) Structure prompts.}
  \label{idea}
\end{figure*}

\begin{figure*}[h]
  \centering
    \includegraphics[scale=0.25]{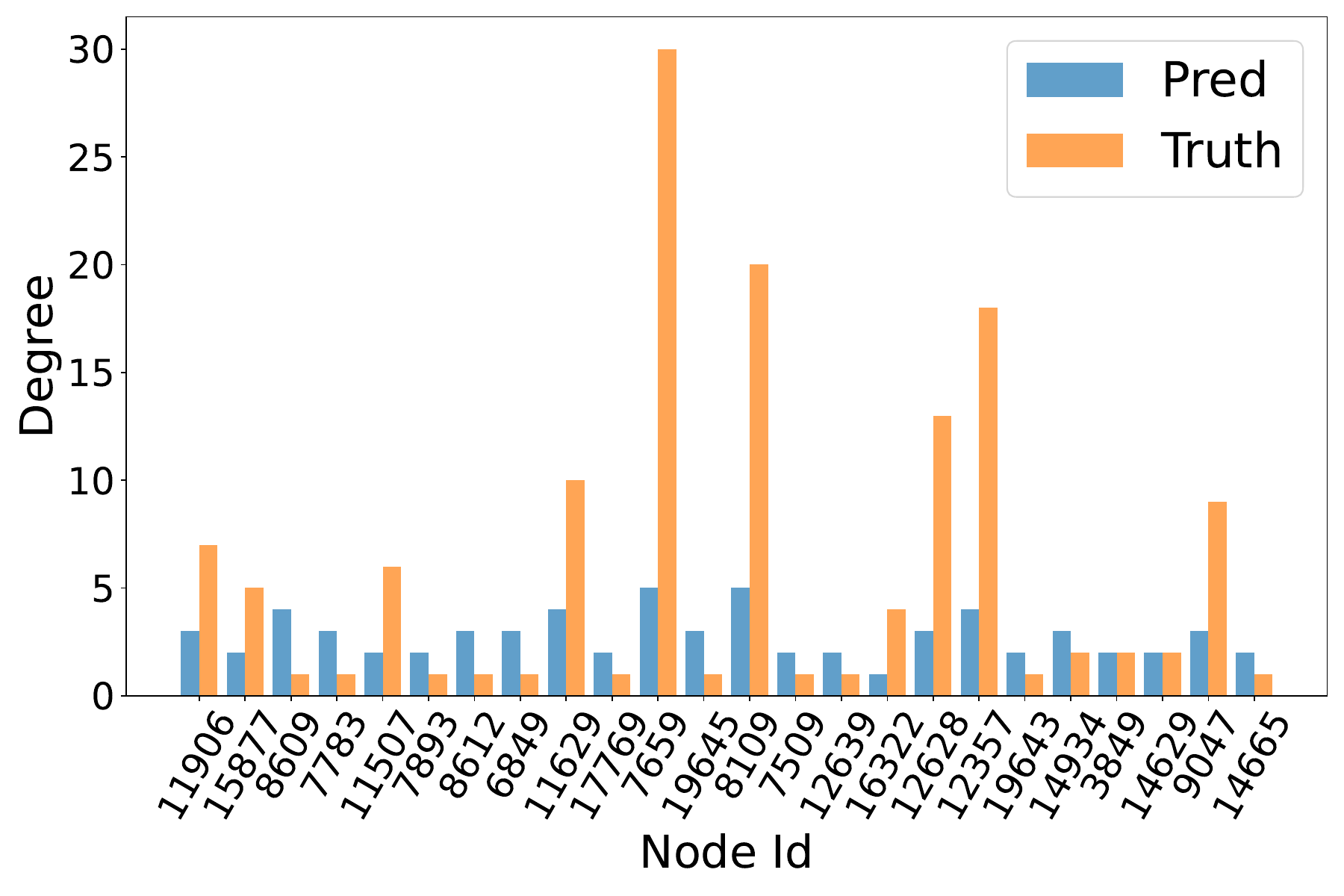}
    \includegraphics[scale=0.25]{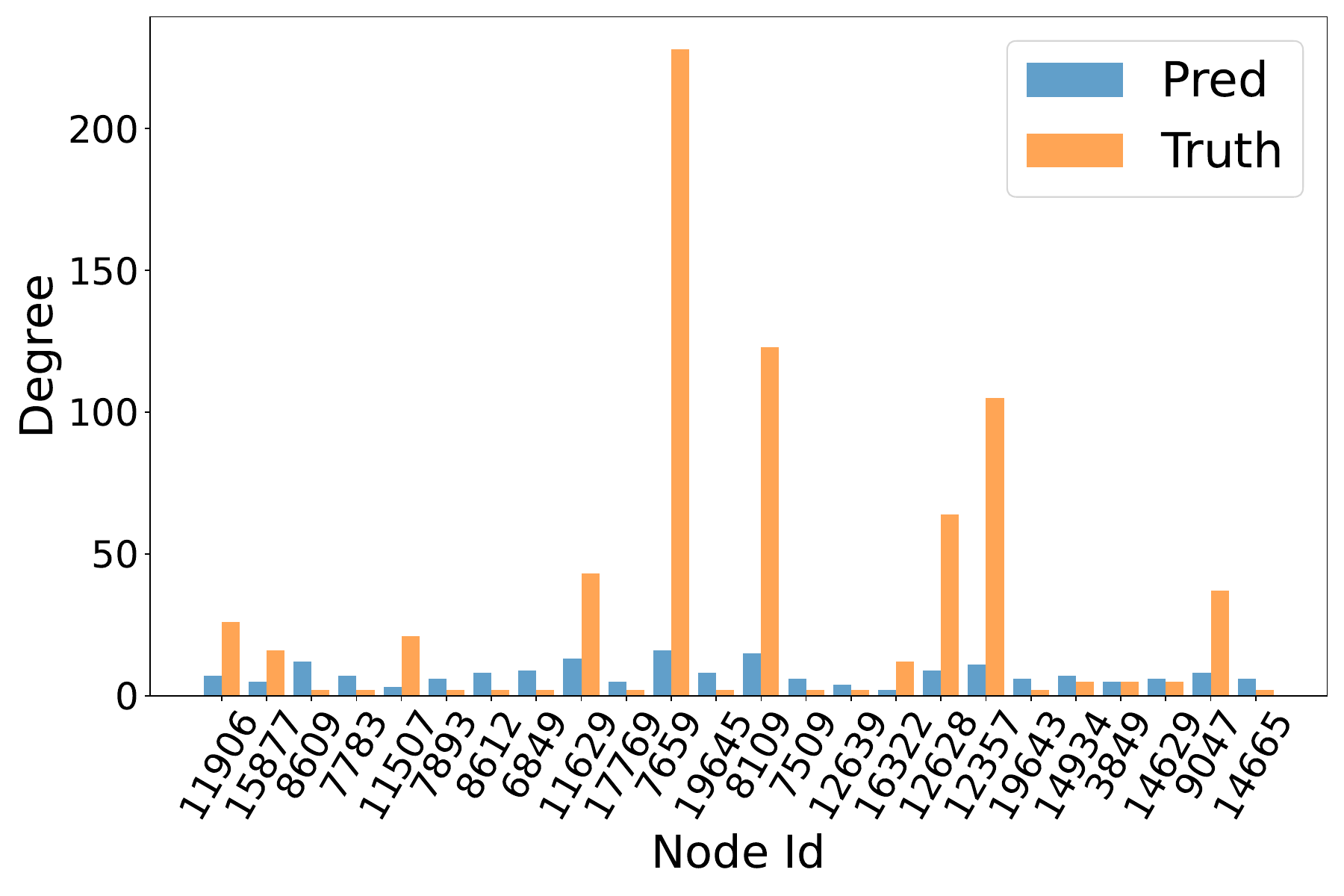}
  \caption{Using node features learned by GCN \cite{gcn2017} on the Pubmed dataset \cite{sen2008} to predict (a) nodes' degree and (b) the sum degree of nodes' neighbors.}
  \label{motivation}
\end{figure*}

Limitations of the representation ability of graph neural networks hinder their performance on node classification, graph classification, and edge prediction tasks \cite{yang2023pmlp, gnnreview2, gnnsurvey2022}. One of the reasons for the unsatisfactory performance is the well-known over-smooth problem \cite{gnnreview3}. Additionally, we found that the expression ability of learned representation fails to capture the intrinsic structural features of graphs, leading to limited performance in downstream tasks. While existing graph neural networks mainly focus on learning features that benefit classifying the nodes (or graphs) into corresponding categories, as shown in Figure~\ref{idea} (b). The target loss of classic graph neural networks is strictly related to downstream tasks,~\textit{e.g.}, node classification, graph classification, and edge prediction. These approaches ignore the intrinsic structure characteristics of graphs during the training process,~\textit{e.g.}, degree of nodes (represented in Figure~\ref{idea} (c)), and distributions of neighbors' degree, one of the neighbor patterns (illustrated in Figure~\ref{idea} (d)). For instance, after being trained for the node classification task on the Pubmed dataset \cite{sen2008}, the GCN \cite{gcn2017} fails to perform well on simple tasks, such as predicting how many nodes are connected with a node and the sum of neighbors' degrees, as shown in Figure~\ref{motivation}. The sum of neighbors' degrees reflects the degree distribution of neighboring nodes to a certain extent. The phenomenon of GNNs failing to capture the intrinsic graph structure and effectively distinguish the essential structural characteristics of nodes is referred to as the ``graph structure lost'' problem in this paper.

The aforementioned ``graph structure lost'' issue inspires us to design some task-independent graph structure prompts to guide graph structure learning in the training process of GNNs. In this way, structural feature learning of graphs can be enhanced and may benefit downstream tasks. Recently, prompt-based tuning methods \cite{schick2020coling,2021graphprompt,promptreview2021,softprompt2021} in natural language processing (NLP) represent a powerful performance by bridging gaps between the pre-training stage and downstream task fine-tuning stages. For instance, given an input sentence of ``The story of the movie is well arranged.'', one can build a prompt template,~\textit{e.g.},  ``I $\_\_$ the movie.'' to facilitate pre-training NLP models. The answer can be chosen from ``like, dislike, ...''.
Motivated by prompt-based methods and multi-task learning, task-independent losses, including first-order and second-order graph structure losses, are proposed in this paper to augment graph structure learning in training GNNs. For example, as shown in Fig \ref{idea} (e), after obtaining the node representation of the orange node E, we design some graph structure prompts and encourage the model to predict ``The degree of E'' is ``\underline{one}'' and the neighbor pattern (e.g., degree distribution) of E during model training. The first-order graph structure loss considers the structure features of the nodes themselves. The second-order graph structure loss comprises the neighboring structural features (neighboring patterns). The computationally intensive higher-order graph structure losses are promising fields to be explored in future works, as the first- and second-order graph structure losses already achieve notable performance. The graph structure prompts in losses are similar to the text prompts in NLP that enable graph neural networks to guide the model to know more about the basic graph structure and learn a better structural representation.  We called this training method as \textbf{G}raph structure \textbf{P}rompt \textbf{L}earning (GPL).

Note that the prompt mechanism was initially used to reduce the gaps between pretraining and inferences in natural language models \cite{promptreview2021} or a transfer learning paradigm to unify pre-training and downstream tasks for GNNs \cite{gppt,graphprompt}. In contrast, we employ graph prompts to train graph neural networks directly without pretraining and relying on extra graph datasets. GPL facilitates the graph neural networks to learn the intrinsic graph structures and work more efficiently in downstream tasks, such as node classification, graph classification, edge prediction,~\textit{etc}. Few works can significantly increase the performance of a wide range of classic graph neural networks without highly improving computation complexity and using extra datasets. Our main contributions are three-fold:

1) We identify a new problem that limits the performance of classic graph neural networks: the lack of intrinsic graph structure learning during training. This is the pioneering work to identify the problem and solve it by involving graph structure prompts.

2) A novel training method, graph structure prompt learning, is proposed to train graph neural networks with graph structure-aware and task-independent losses. With GPL, there is no need to add extra graph data. The effective strategy significantly improves the potential of advanced GNNs and allows them to achieve new state-of-the-art performances.

3) Extensive experiments on eleven real-world datasets demonstrate the effectiveness of GPL. The accuracy of GNNs is improved up to 10.28\%, 16.5\%, and 24.15\% in node classification, graph classification, and edge prediction tasks, respectively.

\section{Related Works}

\textbf{Graph Neural Networks.} 
Graph neural networks have been widely applied in many applications \cite{gnnreview1}. Based on graph spectral theory, ChebNet \cite{gnn2016} and GCN \cite{gcn2017} represent superior learning performance to convolutional neural networks on graph format datasets. Hamilton~\textit{et al.}~\cite{sage2017} introduced a general inductive framework GraphSAGE that leverages neighboring node feature information to generate node representation for unseen data. GAT \cite{gat2018} applies a self-attention mechanism when calculating the weights between neighboring nodes to improve the performance of GCN \cite{gcn2017}.  LightGCN \cite{lgconv2020} accelerates the computations by a light-weight framework. UniMP \cite{transformconv2021} unifies incorporate feature and label propagation at both training and inference time. Wang~\textit{et al.}~\cite{2022location_aware} proposed the LCNN to address the challenge of CNN in constructing receptive fields in graph classification, which learns the location of each node based on its embedding, demonstrating effective task-oriented pattern learning. ARMA \cite{ARMA2021} provides a better global graph structure representation by auto-regressive and moving average filters to provide a more flexible frequency. MM-GNN \cite{bi2023mm}calculates the multi-order moments of the neighbors for each node as signatures, and then uses an element-wise attention-based moment adaptor to assign larger weights to important moments for each node and update node representations. IEA-GNN \cite{zhang2023iea}constructs the candidate anchor point set according to the information entropy of the node, and then obtains the feature information of the node based on the nonlinear distance-weighted aggregate learning strategy of the anchor points of the candidate set, fuses the global position information into the node representation with the selected anchor points. For graph classification, Hu~\textit{et al.}~\cite{gin2018} verify that the graph representation ability of the current graph neural networks is as powerful as the Weisfeiler-Lehman graph isomorphism test. SortPool \cite{sortpool2018} sorts the node features in ascending order along the feature dimension and selects the features of the sorted top-k node to remove influences of unrelated nodes. DiffPool \cite{diffpool2019} applies a differentiable graph pooling module that generates hierarchical representations of graphs. SAGPool \cite{sagpool2019} leverages a self-attention graph pooling layer based on hierarchical graph pooling, which can learn hierarchical representation in an end-to-end manner with relatively few parameters. Recent advanced graph neural networks also include EdgePool \cite{edgepool2019}, TopKPool \cite{topkpool2019},  ASAPool \cite{asapool2020}, MEWISPool \cite{MEWISPool2021}, GPS \cite{GPS2022}. The well-known over-smooth problem prevents the layers of GNNs from going deeper. To solve the problem, DeepGNN \cite{deepgcn2019} applies residual networks to increase the depth of graph neural networks. The PPNP \cite{ppnp2019} decouples dimension transformation and feature propagation. In addition to the over-smooth problem, we found there is another important reason that constraints the expression ability of GNNs, no matter deep or shallow layers. Solving the problem offers great potential for node and graph representation learning. 

\textbf{Graph Pretraining.}
Many new GNNs in recent years \cite{gnnreview3} design the model to address the current issues. However, fewer works focus on proposing a general strategy to improve the training of GNNs. Due to the expensive labeling works of graphs, the benchmark graph datasets are mostly on small scales, which leads to the over-fitting of neural networks. Some works leverage extra unlabeled data to facilitate graph representational learning by developing graph pretraining technologies, including Hu~\textit{et al.}~\cite{hu2019strategies}, GCC \cite{gcc2020}, GPT-GNN \cite{gptgnn2020}, L2P-GNN \cite{L2P2021}, SUGAR \cite{sugar2021}, and CasANGCL \cite{2023casangcl}. Xia~\textit{et al.}~\cite{2023molebert} found the limitations of current AttrMask-based pre-training tasks for molecular prediction, and proposed a novel atom-level pre-training method called Masked Atoms Modeling (MAM). These works suggest that the pretraining and fine-tuning paradigm is a potential way for graph representation learning. 

\textbf{Prompt-Based Learning.}
In recent two years, prompt-based learning methods \cite{schick2020coling,2021graphprompt,softprompt2021} have shown strong capabilities to improve the downstream tasks of NLP. Schick~\textit{et al.}~\cite{schick2020coling} utilize prompt templates to provide label hints of models for classification and generation language tasks. Liu~\textit{et al.}~\cite{softprompt2021} propose prefix-based continuous prompts that support automated template learning in low resources tasks. More works about prompt mechanisms are referred to in the review \cite{promptreview2021}. To reduce the gap between the pretraining objective and the downstream task objective, the GPPT \cite{gppt} employs a masked edge prediction task to pre-train the GNNs and reconstructs the downstream node classification task into a link prediction task. However, the major drawback of graph pretraining methods is that they need ingeniously designed structure patterns that preserve invariant properties of graphs as well as extra graph datasets and expensive training resources. The performance of most current graph pretraining methods relies heavily on certain tasks and domains, not as general and powerful as that in natural language processing \cite{bert2019} since the nodes and edges have diverse meanings in different graphs and domains. Motivated by prompt learning and multi-task learning, we design graph structure prompts that benefit the training of GNNs.

\section{Preliminaries}
Most current graph neural networks follow the message-passing framework \cite{mpnn2017}. They can be described as followings:

A general graph is represented as $G=(V, E)$, where $V$ denotes the node set, and $E$ denotes the edge set.

In node classification, every node $v \in V$ has a corresponding label $y_v$. The objective is to learn a representation vector $h_v$ for each node, such that the node's label $y_v$ can be predicted as $f(h_v)$. Message-passing aggregations update a node's representation by combining the representations of its neighbors. By iterating this process $k$ times, a central node's representation captures the structural information within its neighborhood subgraph. This process can be expressed as:
\begin{equation}
    h_v^{(k)}=COMB^{(k)}\{AGG^{(k)}(h_u^{(k-1)}:u \in N(v)),\ a_v^{(k)}\},
\end{equation}
where $h_v^{(k)}$ is the feature of $v$ in the $k^{(th)}$ iteration and $N(v)$ is the neighbors of $v$. $COMB$ and $AGG$ are the combination and aggregation functions, respectively.

For the graph classification, the readout function aggregates node features during the final iteration to generate a representation $h_G$ of the entire graph. The general format can be expressed as:
\begin{equation}
    h_G=READOUT(h_v^{(k)}|v \in G),
\end{equation}
the READOUT can be a sum, average, max, and other complicated pooling operations.

GCN \cite{gcn2017} is a common aggregation based on message-passing, the layer is computed as:
\begin{equation}
    H^{(l+1)}=\sigma(\hat{A}H^{(l)}W^{(l)}),
\end{equation}
where $\hat{A}$ denotes the normalized adjacency matrix with self-loop. GCN can also be used to apply for the graph classification with a global mean pooling. The equation is:
\begin{equation}
    r_i=\frac{1}{Ni} \sum_{n=1}^{N_i}x_n,
\end{equation}
where $r_i$ is the graph representation, $x$ is the result of aggregated features, and $N_i$ is global nodes.

Message-passing-based aggregations rely on gathering information from node features and their neighbors. This method is insufficient in learning graph structural information from a broad view.  We believe that building proper graph structure prompts facilitates the training of graph neural networks.

\section{Proposed Method}
\subsection{Problem Statements}
Formally, a graph is denoted as $G = (V, E)$. $E$ is presented by an adjacency matrix $A \in \mathbb{R}^{N\times N}$, $N$ is the number of nodes in a graph. The initial node features are represented as $X \in \mathbb{R}^{N\times F}$. The $v_i=X_i$ is the input feature vector of the node $i$ and $F$ is the dimension size.

In the node classification task, the goal of graph neural networks is to learn a model $f_n$ that maps nodes' features $X$ into a vector with a new dimension $X' \in \mathbb{R}^{N\times M_n}$, the $M_n$ is the length of learned features (hidden size of the output layer) by model $f_n$. 
The $v^{f_n}_i = X'_i$ is the output vector of node $i$ by model $f_n$ as $v^{f_n}_i = f_n(v_i, A, X)$. The model $f_n$ contains the graph kernels and aggregation functions of graph neural networks. The probability distribution of class labels is calculated as:
\begin{equation}
    p(\hat{y^i_n}| A, X) = \text{Softmax}(W_n v^{f_n}_i )),
\end{equation}
where $W_n \in \mathbb{R}^{M_n*C_n}$, $M_n$ is the output dimension and $C_n$ is the number of node classes.
The goal of node classification is cross-entropy, denoted as:
\begin{equation}
    \mathcal{L}_n = \sum_{X,Y_n} \sum_i{\log P(\hat{y^i_n}=y^i_n|A,X)},
\label{n_tradtion}
\end{equation}
where $\hat{y_n}$ and $y_n$ are the predicted and actual labels of nodes, respectively.

Graph classification is another basic task of graph neural networks. The model learns a feature vector for each graph. The loss of graph classification task is also cross-entropy, which is expressed as:
\begin{equation}
\begin{aligned}
    p(\hat{y^i_g}| A, X ) &= \text{Softmax}(W_g ( f_{pool} \{f_g(A, X)\} ), \\
    \mathcal{L}_g &= \sum_{X,Y_g} \sum_i{\log P(\hat{y^i_g}=y^i_g|A,X)},
    \end{aligned}
\end{equation}
where $f_g$ is the graph neural network for graph classification. $f_g$ can be the same as $f_n$ or designed differently. $X^g$ is the output features of $f_g$. In some models, e.g., GIN \cite{gin2018}, the $X^g$ is the sum of features from multiple layers, similar to residual networks. The $f_{pool}$ is the readout function or pooling function of graph neural networks. $W_g\in \mathbb{R}^{M_g\times C_g} $ is the output dimension of graph neural networks and $C_g$ is the number of graph classes. $\hat{y_g}$ and $y_g$ are the predicted and actual graph labels.

\subsection{Graph Structure Prompt Learning}

Graph structure prompt learning is not used to pre-train graph neural networks. Instead, it applies task-independent graph structure losses to force the graph neural networks to learn better representation vectors that contain rich intrinsic structure information of graphs. 

\subsubsection{\textbf{Node Classification.}}
In graph neural networks, the feature aggregation function is written as:
\begin{equation}
 X^l = \text{Aggregate}(W^{l-1}X^{l-1}),
  \label{aggregate}
\end{equation}
where $W$ is the weight of $l$th neural networks, $X^l$ is the node feature matrix after $l$ layers that depend on the special implementation of different graph neural networks. $L$ is the set number of graph convolution layers. Specifically, in GCN \cite{gcn2017}, we have:
\begin{equation}
    X^l_{gcn} = \sigma (\hat{D}^{-1/2} \hat{A} \hat{D}^{-1/2} X^{l-1} W^{l-1}_{gcn}),
\end{equation}
where $\hat{A} = A + I$, $X^{0} = X$. $\sigma$ is the activation function (set to Relu in this paper). We apply two graph convolution layers (L=2) and residual networks to combine the input and output features from the last graph convolution layer. The final node representations are calculated by: $X'_{gcn} = X^L_{gcn}+X W^r_{gcn}$, where $W^r \in \mathbb{R}^{F\times M_n}$.

In GCN \cite{gcn2017}, the aggregate function can be seen as the average of neighboring vectors of nodes. In traditional training methods, the output feature $X'_{gcn}$ is taken into Equation \ref{n_tradtion} for model training. Other graph neural networks in experiments follow a similar modification approach.

The first-order graph structure loss is defined as:
\begin{equation}
\mathcal{L}_{1st} = \frac{1}{N} \sum_i (W_1 v_i - \log_2(d_i+ \epsilon))^2,
\end{equation}
where $W_1 \in \mathbb{R}^{M\times 1}$, the parameters of $\mathcal{L}_{1st}$. the $M$ is $M_n$ and $M_g$ in node and graph classification tasks, respectively. $\epsilon$ is set to 1.0 in this paper, $d_i$ is the degree of node $i$.

The second-order graph structure loss is defined as:
\begin{equation}
\mathcal{L}^n_{2nd} = \frac{1}{N\times K} \sum_{i \in V}  \sum_K ( \log_2(D_i+\epsilon) - W_2 v_i )^2 ,
\end{equation}
where $W_2 \in \mathbb{R}^{M\times K}$, the parameters of $\mathcal{L}_{2nd}$, the $M$ can be $M_n$. N=$\|V\|$. $K$ is the maximum number of neighbors in a graph. If the node has fewer neighbors than $K$, the data is padded by zeros to keep the same feature length.  $D_i$ is the logarithm neighboring degree distribution of node $i$. For example, a vector $[1, 0, 0, 3, \cdots]$ means the node $i$ has one neighbor of degree one and four neighbors of degree three. 

For the node classification task, the final loss is: 

\begin{equation}
    \mathcal{L}'_n = \mathcal{L}_n +  \mathcal{L}_{1st} +  \mathcal{L}^n_{2nd},
    \label{n_gpl}
\end{equation}

\subsubsection{\textbf{Graph Classification.}}
For the graph classification task, the readout or pooling mechanism is denoted as:
\begin{equation}
h_g = f_{pool}(\{X^l_v, v \in V\}),
\end{equation}
where $l$ can be $L$ or in a set $\{0,1,...,L\}$. The pooling mechanism or readout function $f_{pool}$ is different in GNNs. For example, GIN \cite{gin2018} aggregates all nodes' features at the same layer as:
\begin{equation}
    h^{l}_{v} = \text{MLP}^{l} \left(  h^{l-1}_v + \sum_{u \in \mathcal{N}(v)} h^{l-1}_u \right),
\end{equation}
where, $h^{l}_{v}$ is the feature vector of a node $v$ in $l$th layer. We apply sum of $h^{l}_{v}$ in GIN \cite{gin2018} before pooling layer as the input vector $v_i$ of $\mathcal{L}_{1st}$ .

The second-order graph structure loss for graph classification is slightly different from $\mathcal{L}^n_{2nd}$ by considering the overall degree distribution of a graph, as the equation:
\begin{equation}
\begin{split}
\mathcal{L}^g_{2nd} = \frac{1}{K} (\frac{1}{N} \sum_{i \in V}  \sum_K ( \log_2(D_i+\epsilon) - W_2 v_i )^2 \\
+\sum_K ( \log_2(D_g+\epsilon) - W^g_2h_g)^2)
\end{split}
\end{equation}

The final loss for graph classification is:
\begin{equation}
    \mathcal{L}'_g = \mathcal{L}_g +  \mathcal{L}_{1st} +\mathcal{L}^g_{2nd},
    \label{g_gpl}
\end{equation}

In this paper, we only apply the first-order and second-order graph structure losses, which are enough to achieve a notable performance.  In the node classification task, GPL applies $\mathcal{L}'_n$ and $\mathcal{L}'_g$ to train models in the node and graph classification task, respectively.

\section{Experimental Analysis}

\subsection{Datasets}

We verify the performance of GPL on classic graph datasets, several representative graph datasets are chosen for each task.

For the node classification and edge prediction tasks, the datasets are as follows:

\textbf{Cora} \cite{sen2008}: It includes 2708 scientific publications on machine learning, and nodes are divided into seven categories. Edges are the citations between papers.

\textbf{Citeseer} \cite{sen2008}: It includes 3327 scientific publications, and nodes are divided into six categories.

\textbf{Pubmed} \cite{sen2008}: It includes 19717 scientific publications on diabetes from the Pubmed database, and nodes are divided into three categories.

\textbf{Photo}: The Amazon Photo network from \cite{2018pitfall}. Nodes represent goods and edges represent that two goods are frequently bought together.

\textbf{Computers}: The Amazon Computers network from \cite{2018pitfall} with same node and edge formats as the Photo. 

\textbf{DBLP} \cite{dblp2017}: The DBLP is a heterogeneous graph containing four types of entities - authors (4,057 nodes), papers (14,328 nodes), terms (7,723 nodes), and conferences (20 nodes).

For Cora, Citeseer, and Pubmed, there are two ways of dividing the training set, one is following the setups as GCN \cite{gcn2017} and another is random selecting.\\

For the graph classification task, the following datasets are used:

\textbf{MUTAG} \cite{1991mutag}: The MUTAG is a dataset of chemical molecules and compounds, with atoms representing junctions and bonds representing edges.

\textbf{PROENTIENS} \cite{2005proteins}: Each node is a secondary structure element, and an edge exists if two nodes are adjacent nodes in the amino acid sequence or 3D space.

\textbf{DD} \cite{dd2003}: Nodes are amino acids, and an edge connects two nodes if the distance between them is less than 6 Angstroms.

\textbf{NCI1} and \textbf{NCI109} \cite{NCI2008}: Datasets on chemical molecules and compounds, where nodes represent atoms and edges represent bonds.

The statistics of graphs are summarized at the top of Table \ref{node_class} and Table \ref{graph_class}. 

\begin{table*}[h!]
  \footnotesize
  \caption{Performance on the node classification task.}
  \begin{center}
    \begin{tabular}{ccccccc}
      \toprule
      \textbf{Datasets} & CORA & Citeseer & Photo & Computers & DBLP & Pubmed \\ \hline
      $|V|$  & 2,708 & 3,327 & 7,650 & 13,752 & 17,716 & 19,717 \\  
      $|E|$  & 5,429 & 4,732 & 238,162 & 491,722 & 105,734 & 44,338 \\ 
      \# classes  & 7 & 6 & 8 & 10 & 4 & 3 \\
      Avg.\#clustering & 0.24 & 0.14 & 0.40 & 0.34 & 0.13 & 0.06 \\\hline
      \textbf{GPPT} & 81.71$\pm$0.19  & 70.31$\pm$0.39 & 92.76$\pm$0.26 & 87.83$\pm$0.28 & - & 85.28$\pm$0.12  \\ \hline
      \textbf{ChebNet}   &82.80$\pm$0.69  & 68.18$\pm$0.85   & 85.71$\pm$9.48 &79.09$\pm$4.17 & 85.75$\pm$0.12  &86.33$\pm$0.16 \\
      \textbf{ChebNet+GPL}  &\textbf{84.69$\pm$0.01} &\textbf{73.14$\pm$0.21} &\textbf{93.67$\pm$0.22} & \textbf{85.73$\pm$0.74} & \textbf{86.33$\pm$0.02} & \textbf{87.22$\pm$0.03} \\   \hline
      \textbf{GCN}          & 82.73$\pm$0.01 & 69.58$\pm$0.36 & 93.52$\pm$0.62 & 88.49$\pm$3.29 & 85.17$\pm$0.05 & 86.05$\pm$0.18\\ 
      \textbf{GCN+GPL}           & \textbf{85.05$\pm$0.01} & \textbf{73.50$\pm$0.31} & \textbf{94.01$\pm$0.14} & \textbf{90.67$\pm$0.25} &  \textbf{86.80$\pm$0.01} & \textbf{88.34$\pm$0.21} \\ \hline
      \textbf{GAT}          & 82.49$\pm$0.88 & 70.84$\pm$0.75  & 85.07$\pm$0.25 & 90.12$\pm$0.31 & 85.07$\pm$0.47 & 83.87$\pm$0.30 \\
      \textbf{GAT+GPL}      & \textbf{84.87$\pm$0.01} & \textbf{73.84$\pm$0.07} & \textbf{95.35$\pm$0.14} & \textbf{90.76$\pm$0.17} & \textbf{86.73$\pm$0.23} & \textbf{85.62$\pm$0.35} \\ \hline
      \textbf{GraphSage}    & 82.10$\pm$0.45 & 71.61$\pm$1.08 & 85.96$\pm$10.20 & 83.99$\pm$1.38 & 85.62$\pm$0.26 & 85.57$\pm$0.21 \\ 
      \textbf{GraphSage+GPL} & \textbf{84.40$\pm$0.37} & \textbf{74.32$\pm$0.52} & \textbf{94.56$\pm$0.15} & \textbf{86.35$\pm$0.81} & \textbf{86.07$\pm$0.26} & \textbf{86.60$\pm$0.28} \\ \hline
      \textbf{LightGCN}       & 80.74$\pm$0.41 & 67.35$\pm$0.65 & 91.58$\pm$0.05 & 83.88$\pm$0.01 & 84.44$\pm$0.04 & 82.56$\pm$0.13 \\ 
      \textbf{LightGCN+GPL}      & \textbf{81.47$\pm$0.52} & \textbf{70.71$\pm$0.12} & \textbf{92.48$\pm$0.02} & \textbf{84.28$\pm$0.05} & \textbf{85.55$\pm$0.02} & \textbf{87.67$\pm$0.01} \\ \hline
      \textbf{UniMP }         & 82.71$\pm$0.89 & 71.67$\pm$0.99 & 94.90$\pm$0.37 & 89.16$\pm$1.54 & 85.37$\pm$0.21 & 85.68$\pm$0.26 \\ 
      \textbf{UniMP +GPL}           & \textbf{84.30$\pm$0.76} & \textbf{74.05$\pm$0.40} & \textbf{95.13$\pm$0.62} & \textbf{89.84$\pm$0.67} &  \textbf{86.22$\pm$0.18} & \textbf{86.73$\pm$0.04} \\ \hline
      \textbf{ARMA} & 82.47$\pm$0.75 & 71.13$\pm$1.39 & 90.00$\pm$4.95 & 83.83$\pm$0.58 & 85.81$\pm$0.13 & 86.78$\pm$0.21 \\ 
      \textbf{ARMA+GPL} & \textbf{83.19$\pm$0.03} & \textbf{74.13$\pm$0.61} & \textbf{94.52$\pm$0.19} & \textbf{85.07$\pm$0.10} & \textbf{86.25$\pm$0.06} & \textbf{87.62$\pm$0.05} \\ \hline
      \textbf{FusedGAT} & 82.39$\pm$0.86 & 70.21$\pm$0.81 & 92.69$\pm$0.44 & 91.05$\pm$0.67 & 93.40$\pm$0.20 & 83.65$\pm$0.13 \\ 
      \textbf{FusedGAT+GPL} & \textbf{85.75$\pm$1.16} & \textbf{73.92$\pm$0.64} & \textbf{95.27$\pm$0.30} & \textbf{91.61$\pm$0.31} & \textbf{95.44$\pm$0.23} & \textbf{88.06$\pm$0.21} \\ \hline
      \textbf{ASDGN} & 81.01$\pm$0.89 & 73.31$\pm$0.45 & 92.22$\pm$0.69 & 83.48$\pm$8.47 & 80.39$\pm$0.28 & 88.84$\pm$0.78 \\ 
      \textbf{ASDGN+GPL} & \textbf{82.34$\pm$0.53} & \textbf{74.86$\pm$0.59} & \textbf{93.69$\pm$0.26} & \textbf{88.31$\pm$0.10} & \textbf{81.21$\pm$0.20} & \textbf{89.03$\pm$0.12} \\ \hline
      \textbf{Avg. Improvement}  &\textbf{1.66} & \textbf{3.17} & \textbf{4.11} & \textbf{2.17} & \textbf{1.06} & \textbf{1.95}  \\
      \textbf{Max. Improvement}   &\textbf{2.38} & \textbf{4.96} & \textbf{10.28} & \textbf{6.64} & \textbf{1.66} & \textbf{5.11} \\
      \bottomrule
    \end{tabular}
    \label{node_class}
  \end{center}
\end{table*}

\subsection{Baselines}
For node classification and edge prediction, we have the following strong baselines:\\

\textbf{ChebNet}  \cite{gnn2016}: It generalizes convolutional operation to graph networks and simplifies calculations based on Chebyshev polynomials.

\textbf{GCN}  \cite{gcn2017}: A commonly used graph convolution network that updates features of a node by averaging its neighboring nodes' attributes.

\textbf{GraphSAGE} \cite{sage2017}: The first inductive graph neural network that aggregates node features by sampling neighbor nodes, it predicts graph context and labels using aggregated information.

\textbf{GAT} \cite{gat2018}: Based on GCN \cite{gcn2017} and self-attention mechanism \cite{transformer2017}, the model aggregates neighbor features via multi-attention heads. GAT has achieved SOTA performance on many datasets in the node classification task.

\textbf{LightGCN} \cite{lgconv2020}: LightGCN learns user and item embeddings by linearly propagating features across the user-item interaction graph. 

\textbf{UniMP} \cite{transformconv2021}: A unified message passing model that effectively combines GNNs and label propagation algorithm (LPA) to incorporate feature and label propagation at both training and inference time. It was named as \textbf{TansformerConv} in PyG \cite{pyg2019}.

\textbf{ARMA} \cite{ARMA2021}: A graph convolutional layer by auto-regressive and moving average filters to provide a more flexible frequency response and better global graph structure representation.

\textbf{Fused GAT} \cite{2022fusedgat}: An optimized version of GAT based on the dgNN \cite{2021dgnn} that fuses message passing computation for accelerated execution and lower memory footprint.

\textbf{ASDGN} \cite{2022anti-symmetric}: A framework for stable and non-dissipative DGN design, conceived through the lens of ordinary differential equations.
\\

\textbf{GPPT} \cite{gppt} uses the masked edge prediction task to pre-train the GNN, and reconstructs the downstream node classification task into a link prediction. We compare it as a baseline for the node classification task.\\\\
For graph classification, we consider the following classic graph neural networks:\\

\textbf{GIN} \cite{gin2018}: A simple graph neural framework that has the same powerful discriminative and representational capabilities as the Weisfeiler-Lehman test.

\textbf{SortPool} \cite{sortpool2018}: A pooling layer that sorts the node features in ascending order along the feature dimension and selects the features of the sorted top-k nodes.

\textbf{DiffPool} \cite{diffpool2019}: A differentiable graph pooling module that can generate hierarchical representations of graphs and can be combined with various graph neural network architectures in an end-to-end fashion.

\textbf{EdgePool} \cite{edgepool2019}: A graph pooling layer relying on the notion of edge contraction that learns a localized and sparse hard pooling transform and considers the graph structure without completely removing nodes.

\textbf{TopKPool} \cite{topkpool2019}: A TopK graph pooling layer to calculate the attention weights of nodes in a graph in a weakly-supervised fashion.

\textbf{SAGPool} \cite{sagpool2019}: A self-attention graph pooling layer based on hierarchical graph pooling, which learns hierarchical representations in an end-to-end manner with relatively few parameters.

\textbf{ASAPool} \cite{asapool2020}: A pooling layer utilizes a novel self-attention network with a modified GNN to capture the importance of each node in a given graph.

\textbf{MEWISPool} \cite{MEWISPool2021}: A graph pooling method based on maximizing mutual information between the pooled graph and the input graph. It employs the Shannon capacity of the graph as an inductive bias during the pooling process.

\textbf{GPS} \cite{GPS2022}: It proposes a general, powerful, scalable (GPS) graph Transformer with linear complexity by decoupling the local real-edge aggregation from the fully-connected Transformer.

\subsection{Model Training}

The learning rate is set from \{0.003, 0.005, 0.0075, 0.01, 0.015, 0.02\}. We use warm training \cite{bert2019} with a small learning rate of 0.00025 for 200 epochs in the task of node classification. The training process continues after warm training stops when 50 epochs no longer reduce validation loss.   In practice, different hyperparameters produce relatively stable performance. We simply set the learning rate to 0.003 in most cases. Slightly adjusting hyperparameters produces positive gains in some cases. Some graph neural networks have additional parameters, which we keep the same to produce comparable results. We ran each model ten times and reported the average accuracy and standard deviation. Since the implementation of models is mostly  based on PyG \cite{pyg2019}, the results may differ from the performances in the original papers. However, the performance is compared fairly.

\subsection{Node Classification}

\begin{table}[htbp]
  \footnotesize
  \setlength{\tabcolsep}{5pt}
  \caption{Node classification by fixed train set.}
  \begin{center}
    \begin{tabular}{cccc}
      \toprule
      \textbf{Datasets} & CORA                                      & Citeseer                             & Pubmed   \\ \hline
      \textbf{GPPT} & 81.40$\pm$0.48 & 69.21$\pm$0.48 & 75.07$\pm$0.93 \\\hline
        \textbf{ChebNet} & 78.71$\pm$0.19 & 64.34$\pm$1.72 & 73.27$\pm$3.55 \\ 
      \textbf{ChebNet+GPL} & \textbf{79.90$\pm$0.30} & \textbf{69.68$\pm$1.07} & \textbf{74.58$\pm$2.10}  \\ \hline
      \textbf{GCN}          & 82.00$\pm$0.01 & 68.05$\pm$0.05 & 78.50$\pm$0.20  \\ 
        \textbf{GCN+GPL}  & \textbf{83.90$\pm$0.02} & \textbf{70.46$\pm$0.14} & \textbf{79.20$\pm$0.02}  \\ \hline
      \textbf{GAT}          & 84.70$\pm$0.01 & 66.70$\pm$2.02 & 76.95$\pm$0.05  \\
      \textbf{GAT+GPL}      & \textbf{85.30$\pm$0.04} & \textbf{70.50$\pm$0.20} & \textbf{78.00$\pm$0.05}  \\ \hline
      \textbf{GraphSage}         & 78.46$\pm$1.11 & 64.52$\pm$1.39 & 76.10$\pm$1.11  \\ 
      \textbf{GraphSage+GPL}          & \textbf{79.56$\pm$0.24} & \textbf{70.59$\pm$0.21} & \textbf{77.80$\pm$0.03}  \\ \hline
      \textbf{LightGCN}       & 77.90$\pm$0.73 & 63.43$\pm$0.89 & 75.10$\pm$0.20  \\ 
        \textbf{LightGCN+GPL}      & \textbf{78.27$\pm$0.37} & \textbf{64.30$\pm$0.57} & \textbf{77.69$\pm$0.45}  \\ \hline
      \textbf{UniMP }           & 78.48$\pm$0.22 & 67.10$\pm$2.62 & 75.81$\pm$0.29  \\ 
      \textbf{UniMP +GPL}           & \textbf{79.34$\pm$0.16} & \textbf{70.95$\pm$0.05} & \textbf{76.85$\pm$0.05}  \\ \hline
      \textbf{ARMA} & 77.11$\pm$0.19 & 64.34$\pm$1.91 & 76.71$\pm$1.62 \\ 
      \textbf{ARMA+GPL} & \textbf{78.07$\pm$0.23} & \textbf{70.37$\pm$0.23} & \textbf{78.00$\pm$0.20}  \\ \hline
      \textbf{FusedGAT} & 74.82$\pm$1.52 & 64.04$\pm$1.13 & 71.59$\pm$2.03 \\ 
      \textbf{FusedGAT+GPL} & \textbf{76.55$\pm$1.05} & \textbf{65.29$\pm$1.41} & \textbf{77.46$\pm$0.27}  \\ \hline
      \textbf{ASDGN} & 60.55$\pm$1.90 & 56.08$\pm$2.49 & 72.81$\pm$1.19 \\ 
      \textbf{ASDGN+GPL} & \textbf{63.59$\pm$1.60} & \textbf{59.81$\pm$1.28} & \textbf{74.94$\pm$0.43}  \\ \hline
     \textbf{Avg. Improvement}  & \textbf{1.31}&\textbf{3.71} &\textbf{1.98}  \\
     \textbf{Max. Improvement}  &\textbf{1.90} &\textbf{6.07} &\textbf{2.59} \\
      \bottomrule
    \end{tabular}
    \label{paper_class}
  \end{center}
\end{table}

Table \ref{node_class} represents the results of node classification by randomly dividing datasets into training, valid, and test sets. If the dataset division is fixed as in the papers \cite{gcn2017, gat2018}, the results are represented in Table \ref{paper_class}. From Table \ref{node_class}, the GPL improves the accuracy of GCN from 82.73 to 85.05 on the Cora. The performance of GCN on the Citeseer improves from 69.58 to 73.5 by a gap of around 4.0. The accuracy of GAT on Cora and Citeseer is improved by a gap of 2.4 and 3.0, respectively. Although GCN and GAT were proposed several years ago, they still have stable high performance on many graph datasets. GPL further improves the results of GCN and GAT. From the results in the tables, after being trained by GPL, the graph neural networks produce better results in all cases. Some models achieved new SOTA performances via GPL. The average clustering and number of classes impact the difficulty of tasks and the effects of GPL, but no strict relationships between them exist. The GPPT performs better than most graph neural networks without GPL in Citeseer by fixing the train sets. However, after being trained by GPL, most GNNs outperform GPPT, indicating significant improvements by GPL. The best performance of GPL is constantly better than GPPT in all datasets and it can be easily applied to all the GNNs. 

\subsection{Graph Classification}

\begin{table*}[h!]
  \footnotesize
  \caption{Performance on the graph classification.}
  \begin{center}
    \begin{tabular}{cccccc}            
      \toprule
      \textbf{Datasets} & MUTAG & PROTEINS    & DD     & NCI1   & NCI109    \\ \hline
      \#graphs  & 188 & 1113     & 1178  &   4110 & 4127  \\ 
      Avg.\#nodes & 17.93 & 39.06 & 284.32  &   29.87 &  29.68 \\ 
      Avg.\#edges & 19.79 & 72.82 & 715.66 & 32.30 &32.13\\
      \#classes & 2 & 2 & 2 & 2 & 2 \\   \hline
      \textbf{GIN}           & 78.50$\pm$2.29 & 76.07$\pm$3.07 & 73.45$\pm$2.41   & 81.07$\pm$0.44 & 77.15$\pm$1.59          \\
      \textbf{GIN+GPL}     &\textbf{95.00$\pm$4.47}       &\textbf{80.98$\pm$2.08} & \textbf{74.11$\pm$2.12} & \textbf{81.17$\pm$1.07} & \textbf{78.21$\pm$1.02} \\ \hline
       \textbf{SortPool}        & 90.00$\pm$2.24 & 75.98$\pm$1.76 & 73.70$\pm$2.66 & 74.96$\pm$1.34 & 75.22$\pm$0.99 \\
       \textbf{SortPool+GPL}    & \textbf{96.00$\pm$2.00} & \textbf{76.25$\pm$0.91} & \textbf{75.71$\pm$1.21} & \textbf{75.38$\pm$0.86} & \textbf{75.94$\pm$1.15} \\ \hline
        \textbf{DiffPool} &86.00$\pm$5.39  & 81.61$\pm$2.32 & - &72.41$\pm$1.07&72.27$\pm$1.64\\
      \textbf{DiffPool+GPL} & \textbf{89.00$\pm$3.00} &\textbf{81.96$\pm$2.86} & - &\textbf{73.14$\pm$1.31} &\textbf{73.07$\pm$1.33}
      \\ \hline
      \textbf{EdgePool}          &72.50$\pm$2.50  & 58.40$\pm$2.10 & 58.40$\pm$2.10 & 71.70$\pm$0.49 & 67.31$\pm$1.09\\
      \textbf{EdgePool+GPL} &\textbf{80.50$\pm$1.50}       & \textbf{73.84$\pm$1.33} & \textbf{71.09$\pm$0.94} & \textbf{73.31$\pm$0.81} & \textbf{72.90$\pm$0.77} \\ \hline
      \textbf{TopKPool}          & 70.00$\pm$3.87 & 73.39$\pm$1.25 & 71.93$\pm$1.56 & 70.29$\pm$1.11 & 68.41$\pm$1.58 \\
      \textbf{TopKPool+GPL}      &\textbf{80.50$\pm$5.22}  & \textbf{76.96$\pm$0.78} & \textbf{74.79$\pm$1.35} & \textbf{71.14$\pm$2.37} & \textbf{71.26$\pm$1.21} \\ \hline

      \textbf{SAGPool}          & 68.00$\pm$4.58 & 75.71$\pm$1.08 & 76.47$\pm$3.36 & 71.80$\pm$2.90 & 73.82$\pm$1.78 \\
       \textbf{SAGPool+GPL}     & \textbf{73.00$\pm$8.71} & \textbf{78.21$\pm$2.15} & \textbf{78.82$\pm$3.53} & \textbf{72.60$\pm$1.37} & \textbf{74.93$\pm$2.36} \\ \hline

        \textbf{ASAPool}        & 65.00$\pm$4.47 & 72.32$\pm$2.91 & 62.18$\pm$0.01 & 65.94$\pm$0.57 & 69.49$\pm$0.94\\        
      \textbf{ASAPool+GPL}      & \textbf{79.50$\pm$3.50} &\textbf{76.70$\pm$1.29} &\textbf{70.34$\pm$2.84} & \textbf{66.40$\pm$1.08} & \textbf{69.71$\pm$0.70}   \\  \hline
    
      \textbf{MEWISPool}        & 80.00$\pm$3.16 & 69.40$\pm$1.44 & 69.60$\pm$3.40 & 56.34$\pm$4.86 & 59.00$\pm$7.90\\
      \textbf{MEWISPool+GPL}    & \textbf{89.47$\pm$3.33} & \textbf{70.60$\pm$2.86} & \textbf{71.00$\pm$4.00} & \textbf{67.43$\pm$6.00} & \textbf{67.40$\pm$6.60} \\ \hline
      \textbf{GPS}        & 66.50$\pm$8.08 & 64.46$\pm$4.26 & 53.11$\pm$0.63 & 62.24$\pm$7.66 & 56.98$\pm$5.22\\
      \textbf{GPS+GPL}    & \textbf{76.50$\pm$3.20} & \textbf{67.85$\pm$7.15} & \textbf{58.61$\pm$6.57} & \textbf{63.23$\pm$7.84} & \textbf{65.41$\pm$6.16} \\ \hline
      \textbf{Avg. Improvement}  & \textbf{9.21} &  \textbf{4.00} & \textbf{4.45} & \textbf{1.89} & \textbf{3.24}\\
      \textbf{Max. Improvement}  & \textbf{16.5} & \textbf{15.44} & \textbf{12.69} & \textbf{11.09} & \textbf{8.40} \\
      \bottomrule
    \end{tabular}
    \label{graph_class}
  \end{center}
\end{table*}

In the task of graph classification, as shown in Table \ref{graph_class}, we can see that GPL improves the performance of GIN \cite{gin2018} and SortPool \cite{sortpool2018} is improved from 78.5 to 95.0 and 90.0 to 96.0 on the dataset MUTAG, respectively, which indicates they have a higher performance potential after applying GPL during training. The accuracies of EdgePool \cite{edgepool2019} and ASAPool \cite{asapool2020} on the dataset DD are increased from 58.4 to 71.09 and 62.18 to 70.34, respectively, with significant margins. The performance of MEWISPool and EdgePool \cite{edgepool2019} in NCI109 is improved by a gap of 8.4 and 5.61, respectively. The improvements vary across models and datasets. But GPL consistently improves the performance of all the graph neural networks. Few known methods can be so effective in improving the performance of graph neural networks on the graph classification task.

\begin{table*}[h!]
  \footnotesize
  \caption{Performance of edge prediction.}
  \begin{center}
    \begin{tabular}{ccccccc}
      \toprule
      \textbf{Datasets} & CORA & Citeseer & Photo & Computers & DBLP & Pubmed   \\ \hline  
       \textbf{ChebNet} & 85.34$\pm$1.56 & 80.05$\pm$1.31 & 75.61$\pm$0.05 & 80.31$\pm$3.45 & 91.76$\pm$0.33 & 85.93$\pm$0.41 \\ 
      \textbf{ChebNet+GPL} & \textbf{89.04$\pm$1.32} & \textbf{92.70$\pm$0.85} & \textbf{86.71$\pm$0.77} & \textbf{84.14$\pm$0.75} & \textbf{93.27$\pm$0.26} & \textbf{87.85$\pm$0.32}  \\ \hline
      \textbf{GCN}          & 92.90$\pm$0.76 & 92.98$\pm$0.52 & 92.73$\pm$2.2 & 91.40$\pm$1.26 & 96.67$\pm$0.14 & 95.70$\pm$0.14  \\ 
        \textbf{GCN+GPL}  & \textbf{93.50$\pm$0.9} & \textbf{93.40$\pm$0.73} & \textbf{93.71$\pm$0.89} & \textbf{91.62$\pm$1.56} & \textbf{96.71$\pm$0.2} & \textbf{96.66$\pm$0.27}  \\ \hline
      \textbf{GAT}          & 92.04$\pm$0.84 & 91.97$\pm$0.74 & 94.27$\pm$0.18 & 90.38$\pm$1.10 & 95.83$\pm$0.21 & 93.37$\pm$0.16\\
      \textbf{GAT+GPL}      & \textbf{93.10$\pm$0.87} & \textbf{93.05$\pm$0.25} & \textbf{94.85$\pm$0.36} & \textbf{90.78$\pm$0.25} & \textbf{96.07$\pm$0.21} & \textbf{94.27$\pm$0.18}  \\ \hline
      \textbf{GraphSage}         & 90.69$\pm$1.03 & 87.90$\pm$1.81 & 65.46$\pm$10.40 & 84.92$\pm$2.49 & 94.78$\pm$0.21 & 89.06$\pm$0.04\\ 
      \textbf{GraphSage+GPL}          & \textbf{92.54$\pm$0.94} & \textbf{93.00$\pm$0.75} & \textbf{89.61$\pm$0.44} & \textbf{86.65$\pm$0.32} & \textbf{95.40$\pm$0.24} & \textbf{90.54$\pm$0.16}  \\ \hline
      \textbf{LightGCN}       & 89.49$\pm$1.06 & 84.31$\pm$1.03 & 85.56$\pm$0.42& 89.95$\pm$4.05 & 93.03$\pm$0.19 & 85.56$\pm$0.42\\ 
       \textbf{LightGCN+GPL}      & \textbf{92.13$\pm$0.66} & \textbf{93.19$\pm$0.76} & \textbf{90.93$\pm$0.76} & \textbf{93.25$\pm$1.17} & \textbf{94.98$\pm$0.23} &\textbf{87.66$\pm$0.31}  \\ \hline
      \textbf{UniMP }           & 90.18$\pm$0.45 & 87.81$\pm$1.05 & 85.00$\pm$3.08 & 84.92$\pm$2.49 & 95.05$\pm$0.20 & 87.17$\pm$0.65 \\ 
      \textbf{UniMP +GPL}           & \textbf{91.55$\pm$0.98} & \textbf{93.26$\pm$1.07} &   \textbf{88.69$\pm$1.25} & \textbf{85.52$\pm$0.61} & \textbf{95.48$\pm$0.15} & \textbf{88.37$\pm$0.39}  \\ \hline
      \textbf{ARMA} & 87.23$\pm$1.38 & 81.99$\pm$1.09 & 75.56$\pm$3.85 & 78.13$\pm$1.87 & 92.36$\pm$0.25 & 88.37$\pm$0.44  \\ 
      \textbf{ARMA+GPL} & \textbf{90.95$\pm$1.09} & \textbf{91.44$\pm$0.99} & \textbf{93.78$\pm$0.32} & \textbf{91.50$\pm$0.65} & \textbf{94.96$\pm$0.20} & \textbf{90.04$\pm$0.21 }  \\ \hline
      \textbf{FusedGAT} & 89.41$\pm$1.00 & 82.19$\pm$1.13 & 91.78$\pm$1.45 & 91.16$\pm$0.44 & 86.04$\pm$0.32 & 86.25$\pm$0.39  \\ 
      \textbf{FusedGAT+GPL} & \textbf{92.72$\pm$0.43} & \textbf{92.61$\pm$1.23} & \textbf{93.81$\pm$0.43} & \textbf{91.81$\pm$1.04} & \textbf{88.60$\pm$0.63} & \textbf{88.89$\pm$0.63 }  \\ \hline
      \textbf{ASDGN} & 91.45$\pm$1.23 & 90.36$\pm$0.84 & 87.94$\pm$2.37 & 86.70$\pm$1.61 & 95.30$\pm$0.22 & 94.55$\pm$0.57  \\ 
      \textbf{ASDGN+GPL} & \textbf{92.37$\pm$0.74} & \textbf{93.71$\pm$0.51} & \textbf{90.00$\pm$0.60} & \textbf{89.41$\pm$0.65} & \textbf{96.27$\pm$0.13} & \textbf{95.06$\pm$0.23 }  \\ \hline
     \textbf{Avg. Improvement}  & \textbf{2.13}&\textbf{6.31} &\textbf{7.58} &\textbf{2.97}  & \textbf{1.21} & \textbf{1.48}\\
     \textbf{Max. Improvement}  &\textbf{3.72} &\textbf{12.65} &\textbf{24.15} &\textbf{13.37}  & \textbf{2.60} & \textbf{2.64}\\
      \bottomrule
    \end{tabular}
    \label{edge_prediction}
  \end{center}
\end{table*}

\subsection{Edge Prediction}
Table \ref{edge_prediction} reports the results of edge prediction. The experimental setup refers to PyG repository \cite{pyg2019}, predicting whether an edge exists between two nodes utilizing the learned representations. We use the same datasets and approaches in node classification for the edge prediction task. The datasets are divided randomly. Edges in the datasets are sampled into equal numbers of positive and negative pairs. The AUC score is utilized as the evaluation metric.  From Table~\ref{edge_prediction}, GNNs with GPL achieve much  better results. Especially on the Citeseer and Photo, GPL increases the average performance by 6.15 and 9.16, respectively. The improvement taken by GPL in edge prediction is meritorious.

The results of node classification, graph classification, and edge prediction have shown the stability and effectiveness of GPL in improving the performance of graph neural networks.
There are no apparent underlying connections between degree/neighbor patterns and the downstream tasks. However, the losses facilitate the GNNs ``know'' more about the graphs and lead to better performance.

\subsection{Sensitivity and Ablation Study}

\subsubsection{\textbf{Parameter Sensitivity}}
\begin{figure}[h]
  \centering
  \subfigure[Cora]{
    \includegraphics[scale=0.27]{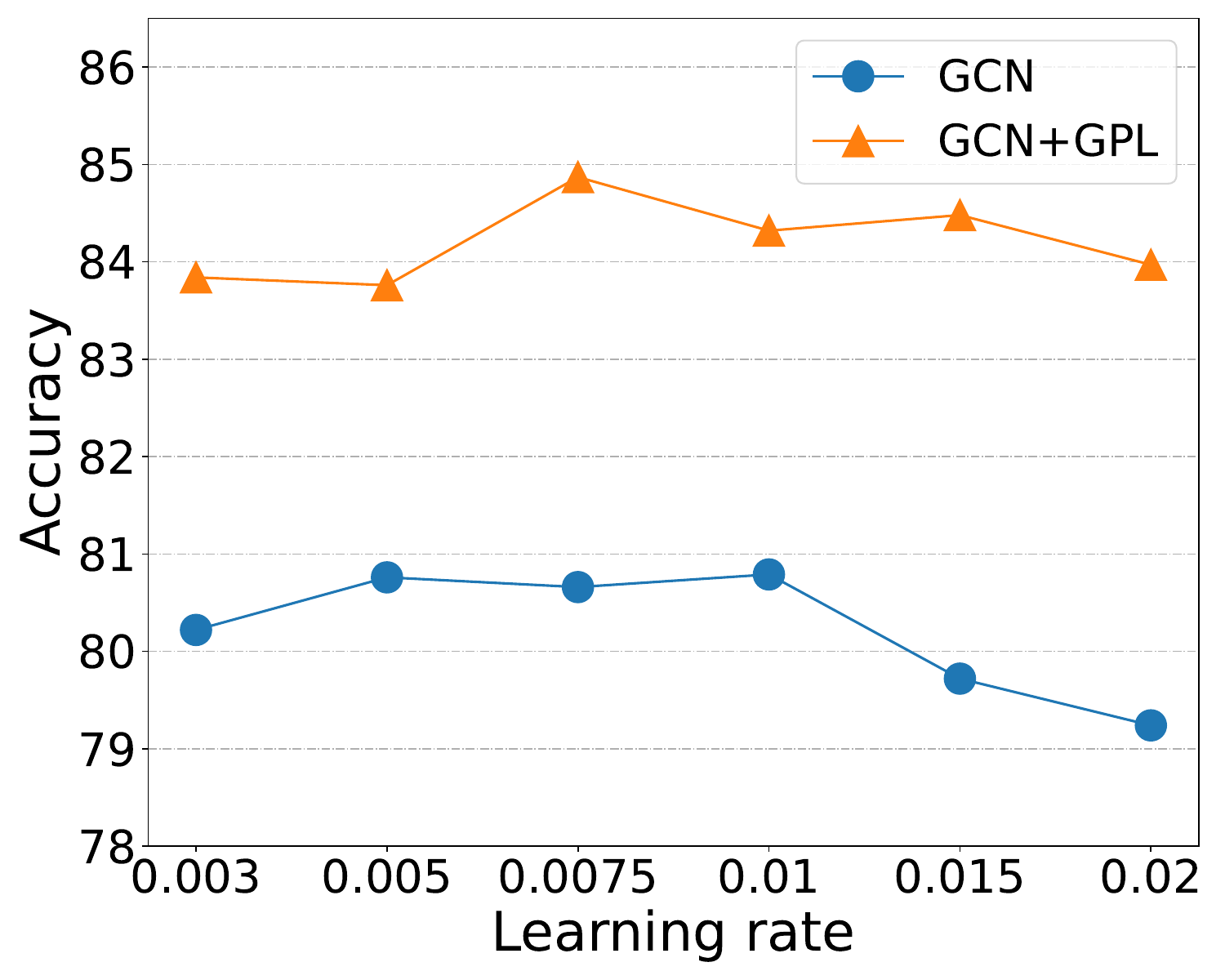}
  }
  \subfigure[Citeseer]{
    \includegraphics[scale=0.27]{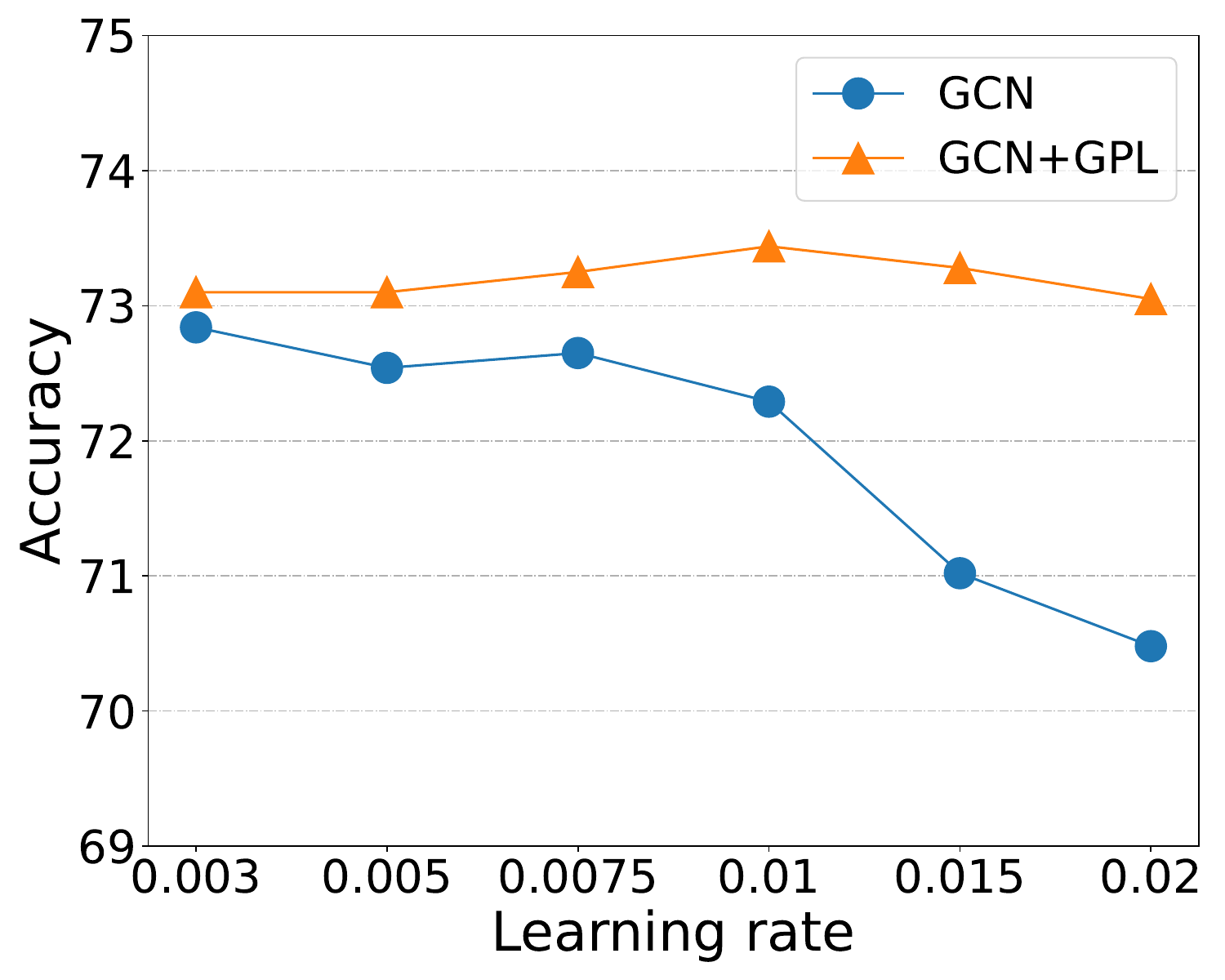}
  }
  \subfigure[Computers]{
    \includegraphics[scale=0.27]{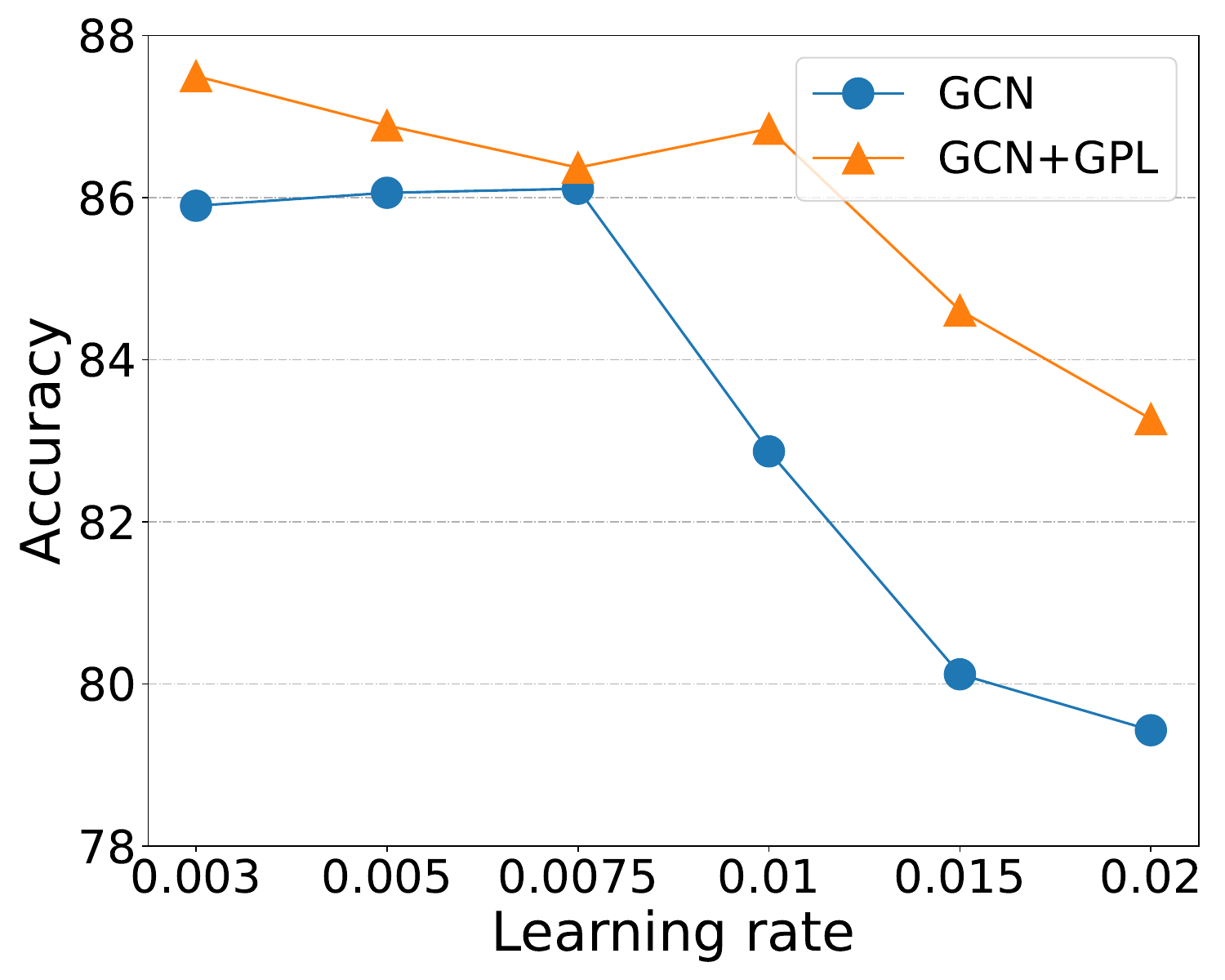}
  }
  \subfigure[DBLP]{
    \includegraphics[scale=0.27]{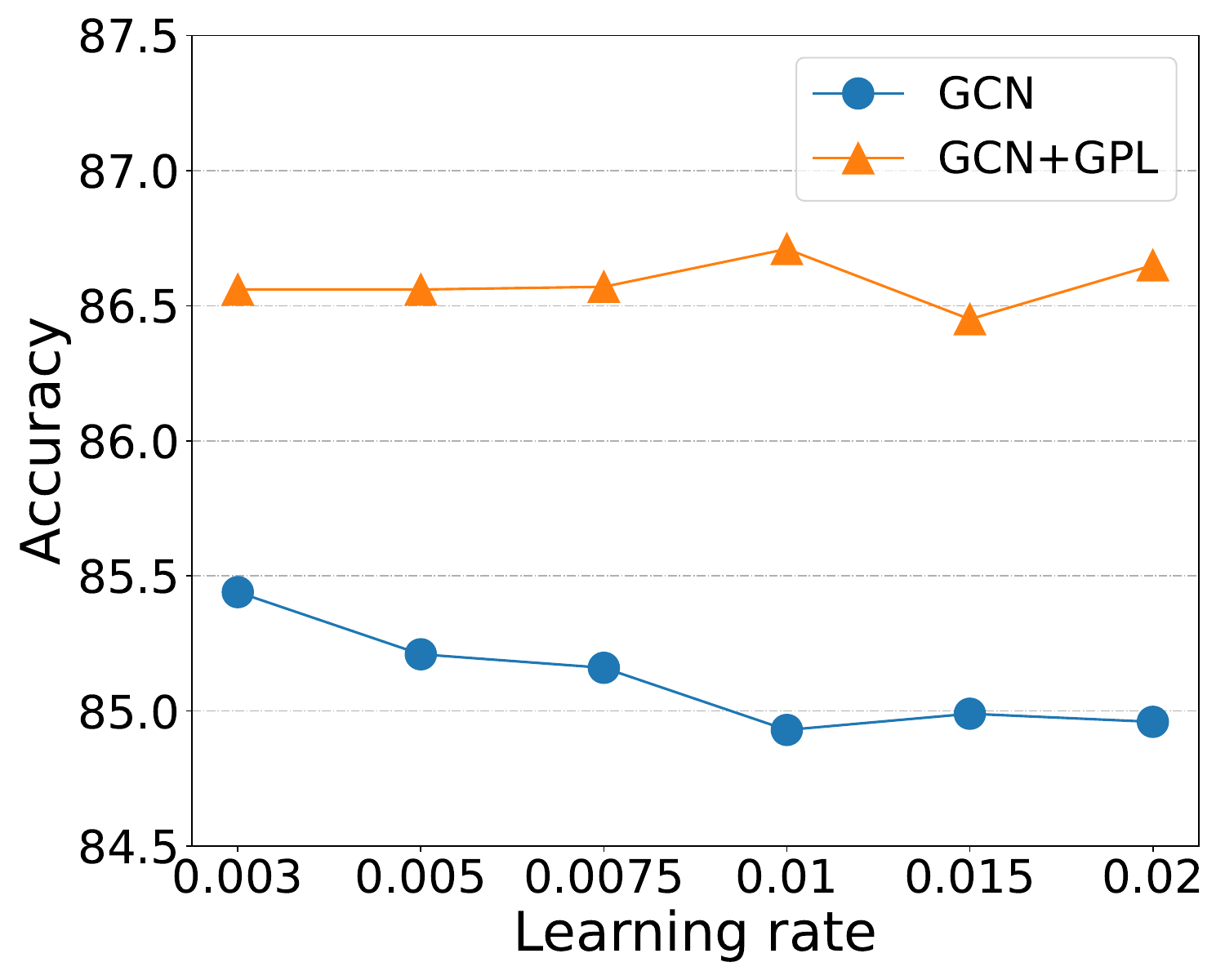}
  }
  \caption{How learning rate impacts the performance of GCN and GCN+GPL.}
  \label{lr}
\end{figure}

We validate the parameter sensitivity of GPL to explore its stability in performance. The learning rate is chosen from \{0.003, 0.005, 0.0075, 0.01, 0.015, 0.02\} by GCN \cite{gcn2017} on the Cora, Citeseer, Computers, and DBLP datasets. The results are shown in Figure~\ref{lr}. For GCN, a smaller learning rate is recommended. There is a trend that the gaps between the two methods widen as the learning rate increases.  Without GPL, GCN's accuracy drops down rapidly when the learning rate is set larger than certain values. However, after being trained by GPL, the performance of GCN is more stable. Similar phenomena occur in other graph neural networks, we take experiments on GCN as an example.

\subsubsection{\textbf{Ablation Study}}

To verify the impacts of components in GPL,  we remove the \textit{residual networks}, $\mathcal{L}_{1st}$, and $\mathcal{L}^n_{2nd}$, respectively, or replace $\mathcal{L}_{1st}$ with $\mathcal{L}_{Centrality}$ using node centrality. We also concatenate features of degree with GNN and mark it as ``GNN+Degree''. We also train GNNs by $\mathcal{L}_n / \mathcal{L}_g $, $\mathcal{L}_{1st}$, and $\mathcal{L}^n_{2nd}$ sequentially, and denote it as GNN+Seq-task. The results are recorded in Table \ref{ablation}. When $\mathcal{L}_{1st}$ or $\mathcal{L}^n_{2nd}$ are removed, the accuracy drops on different datasets. Removing residual networks leads to a larger drop in accuracy. However, the residual networks contribute little to the accuracy of the original GCN and GAT. Simply combing features of degree or some other features of graphs contributes less to GCN and GAT, which might be because static statistic features have to work with complicated graph structures. $\mathcal{L}_{Centrality}$ is another candidate of $\mathcal{L}_{1st}$, but not stable in dataset Cora and Computers. According to the previous work \cite{gcn2017}, GCN also contains node degrees' representation. Part of the information in representations might lose after training for classification tasks. From Table \ref{ablation}, by combing the \textit{residual networks}, $\mathcal{L}_{1st}$ and $\mathcal{L}^n_{2nd}$, the model achieves the best performance in most cases.

\begin{table*}[!ht]
  \footnotesize
  \centering
  \caption{Ablation study on node classification.}
  \begin{center}
    \begin{tabular}{ccccccc}
      \toprule
      \textbf{Datasets} & Cora & Citeseer   & Photo  &  Computers & DBLP & Pubmed              \\ \hline
      \textbf{GCN+Degree}& 82.73 & 70.25 & 93.64 & 89.24 & 84.93 & 85.68\\
      \textbf{GCN} & 82.73 & 69.58 &  93.52 & 88.49 & 85.17 & 86.05    \\
      \textbf{GCN+GPL}      & \textbf{85.05} &\textbf{73.50}      &      \textbf{94.01}   &\textbf{90.67}   &\textbf{86.80} & \textbf{88.34}                                \\
      \textbf{GCN+GPL-$Residual$}   & 83.15 & 71.02   &  93.26      &  90.09  &85.21   & 85.94  \\
        \textbf{GCN+GPL-{$\mathcal{L}_{1st}$}}    & 82.73      & 73.31   & 93.90 & 90.18 & 86.64 & 87.34        \\ 
      \textbf{GCN+GPL-{$\mathcal{L}^n_{2nd}$}}    & 82.38     & 73.30  & 93.95 &  90.30  & 86.67  &    87.24    \\
      
      \textbf{GCN+GPL with $\mathcal{L}_{Centrality}$}& 80.65 & 72.62 & 93.94 & 84.96 & 85.39 & 87.54\\
      \textbf{GCN Seq-task}& 72.90 & 71.53 & 86.44 & 72.90 & 84.56 & 55.46\\\hline
            \textbf{GAT+Degree}& 80.57 & 70.45 & 93.44 & 88.26 & 85.57 & 83.69\\
      \textbf{GAT} & 82.49 & 70.84 & 85.07 & 90.12  & 85.07 & 83.87 \\
      \textbf{GAT+GPL}  & \textbf{84.87} &\textbf{73.84}      & \textbf{95.35}  & \textbf{90.76} &\textbf{86.73}  & \textbf{85.62}  \\
        \textbf{GAT+GPL-$Residual$} & 82.73 &70.63      & 93.84    &  90.44  &86.08    & 85.35                                                                       \\
      \textbf{GAT+GPL-{$\mathcal{L}_{1st}$}}   & 83.62  & 72.75     &    94.99     &     90.34     & 86.50  &  85.60    \\ 
      \textbf{GAT+GPL-{$\mathcal{L}^n_{2nd}$}}    & 83.27   & 73.23    &  \textbf{95.35}  & 90.13    & 86.68  & 85.50  \\ 
      \textbf{GAT+GPL with $\mathcal{L}_{Centrality}$} & 84.30 & 72.75 & 94.98 & 90.61 & 86.50 & 85.59\\
      \textbf{GAT Seq-task}& 83.54 & 72.38 & 94.50 & 88.90 & 85.67 & 75.21\\
      \bottomrule
    \end{tabular}
    \label{ablation}
  \end{center}
\end{table*}

\subsection{Deep-layer Effects}
To verify the effectiveness of GPL on GNNs with deep layers, we modify the GCN+GPL with four, six, and eight layers. Table~\ref{multi_layers} shows the performances of GCN with deeper layers on node classification. As the layers increase, the accuracy of GCN sharply decreases. When GCN in Cora has four, six, and eight layers, the improvement gaps between GCN and GCN+GPL grow from 8.4 to 9.7 to 15.2. The GCN+GPL is much more stable than GCN. The accuracy grows slightly from four to six layers, which offers a new possible direction to address the over-smooth problem.

\begin{table*}[!ht]
    \centering
    \footnotesize
    \caption{Multi-layers of GCN on node classification.}
    \begin{tabular}{ccccccc}
    \toprule
        \textbf{Dataset} & Cora & Citeseer & Photo & Computers & DBLP & Pubmed \\ \hline
        \textbf{GCN(4 layers)} & 74.17 & 70.15 & 90.97 & 81.12 & 84.62 & 85.79 \\
        \textbf{GCN+GPL(4 layers)} & \textbf{82.56} & \textbf{71.88} & \textbf{94.04} & \textbf{86.95} & \textbf{85.13} & \textbf{86.08} \\ \hline
        \textbf{GCN(6 layers)} & 72.50 & 68.99 & 90.29 & 80.25 & 84.43 & 85.72 \\
        \textbf{GCN+GPL(6 layers)} & \textbf{82.23} & \textbf{72.05} & \textbf{94.35} & \textbf{87.71} & \textbf{85.39} & \textbf{86.17} \\ \hline
        \textbf{GCN(8 layers)} & 67.03 & 58.98 & 81.55 & 64.98 & 84.14 & 63.97 \\
        \textbf{GCN+GPL(8 layers)} & \textbf{82.21} & \textbf{70.39} & \textbf{94.08} & \textbf{86.87} & \textbf{84.76} & \textbf{85.78} \\
        \bottomrule
    \end{tabular}
    \label{multi_layers}
\end{table*}

\subsection{Representation Visualizations} 
\begin{figure}[h!]
  \centering
  \subfigure[After GCN]{
    \includegraphics[scale=0.25]{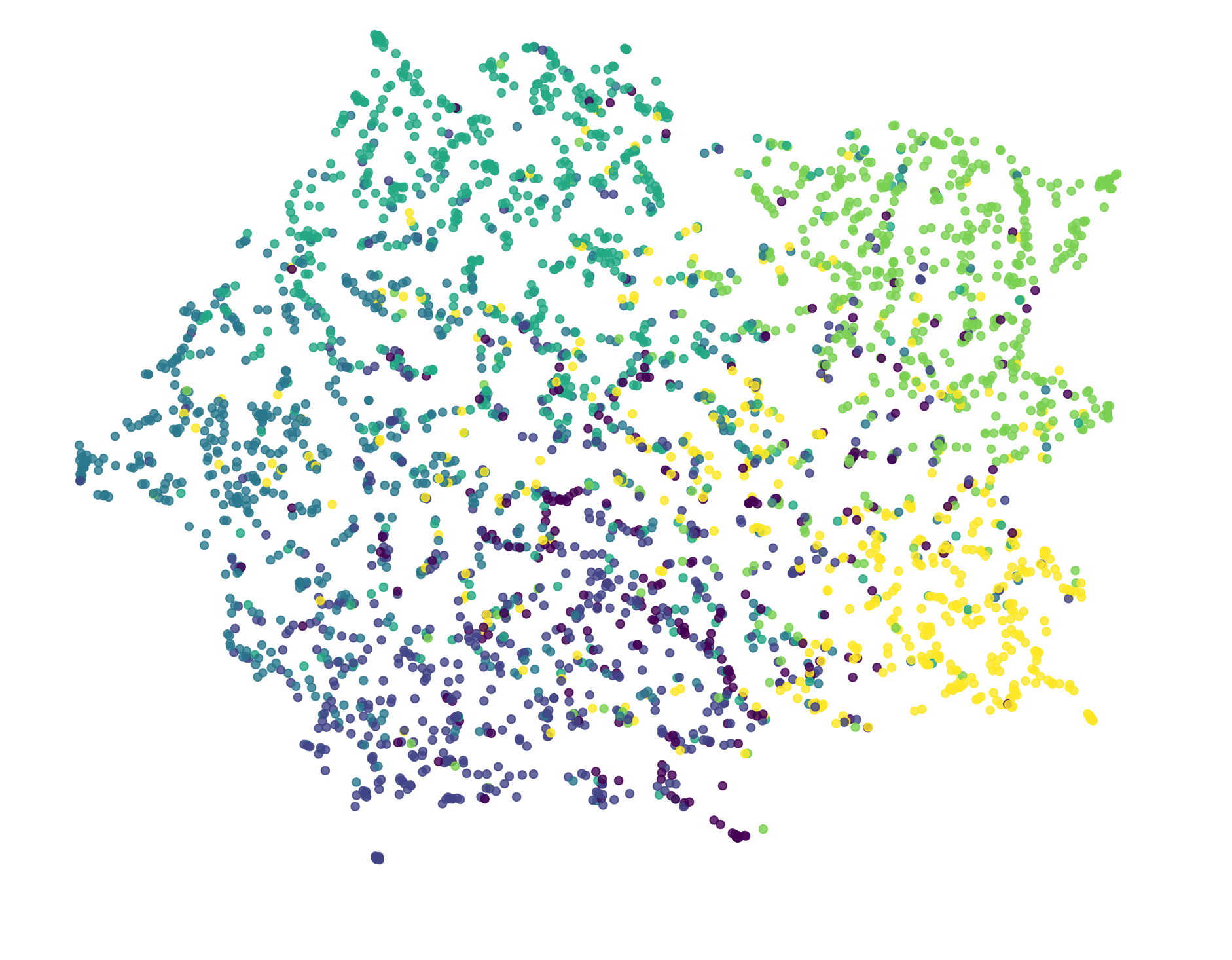}
  }
   \subfigure[After GCN+GPL]{
     \includegraphics[scale=0.25]{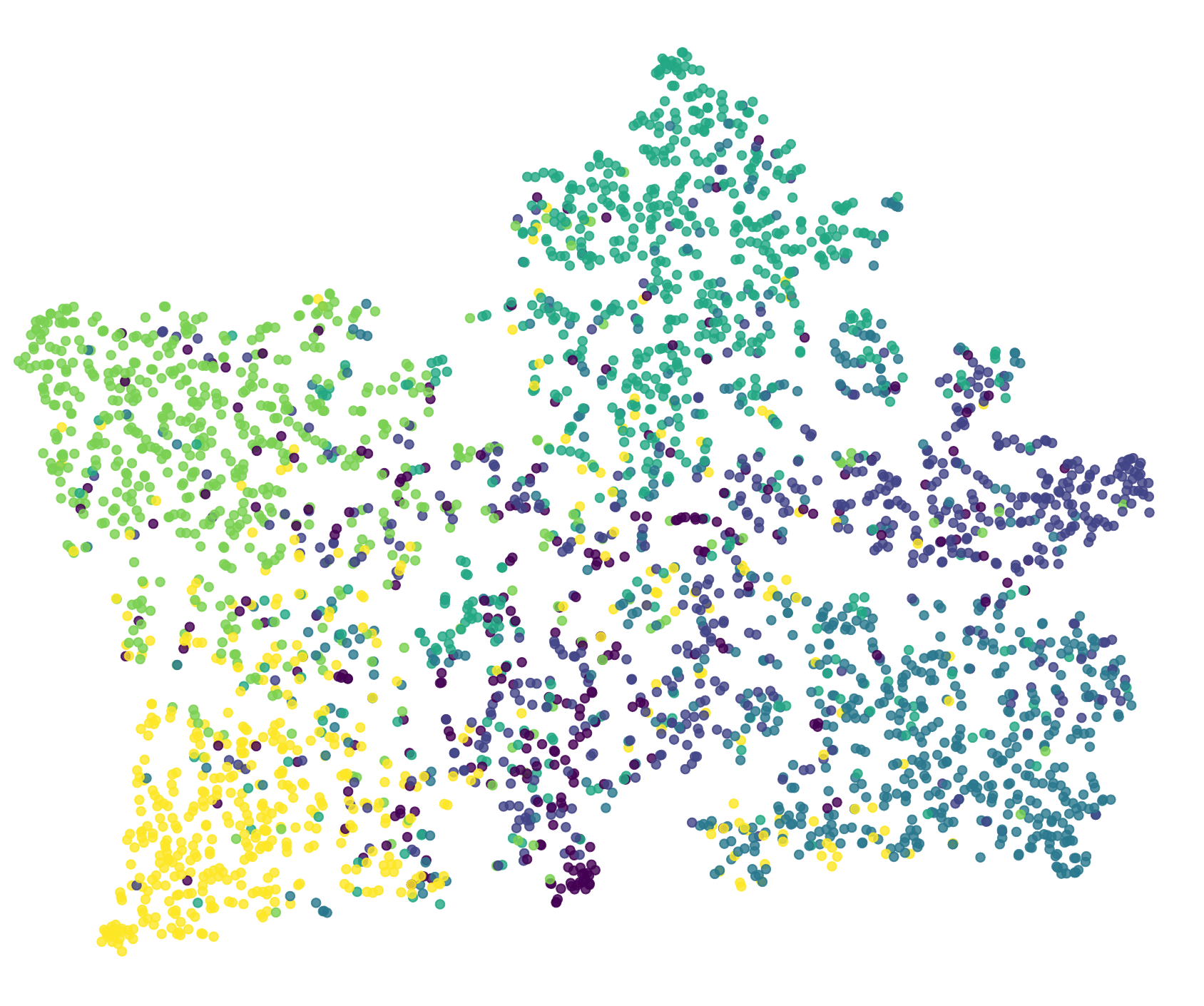}
   }
  \caption{Visualization of nodes' representation after training on Cora.}
  \label{nc_vis}
\end{figure}

\begin{figure}[h!]
  \centering
  \subfigure[After SAGPool]{
    \includegraphics[scale=0.25]{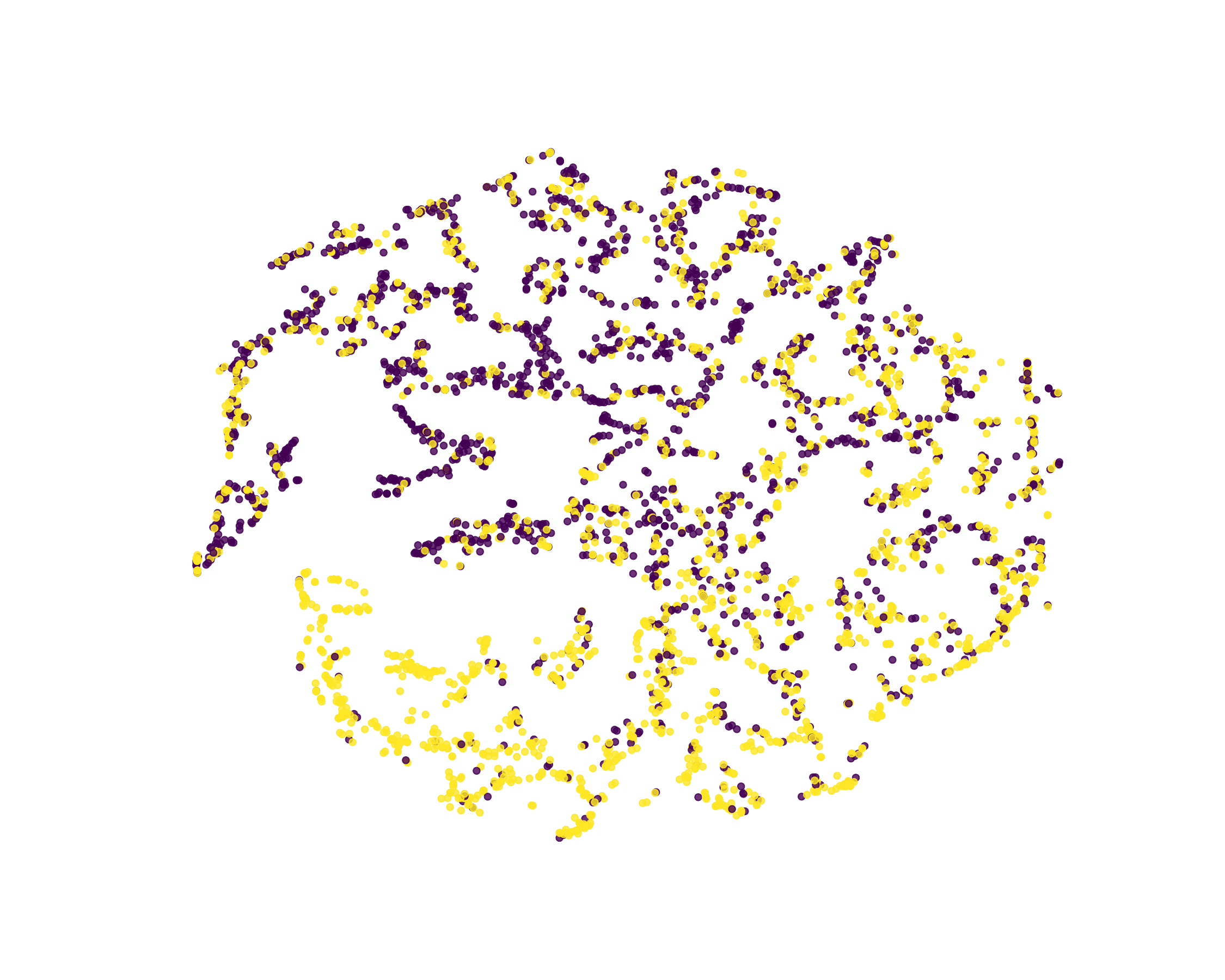}
  }
   \subfigure[After SAGPool+GPL]{
     \includegraphics[scale=0.25]{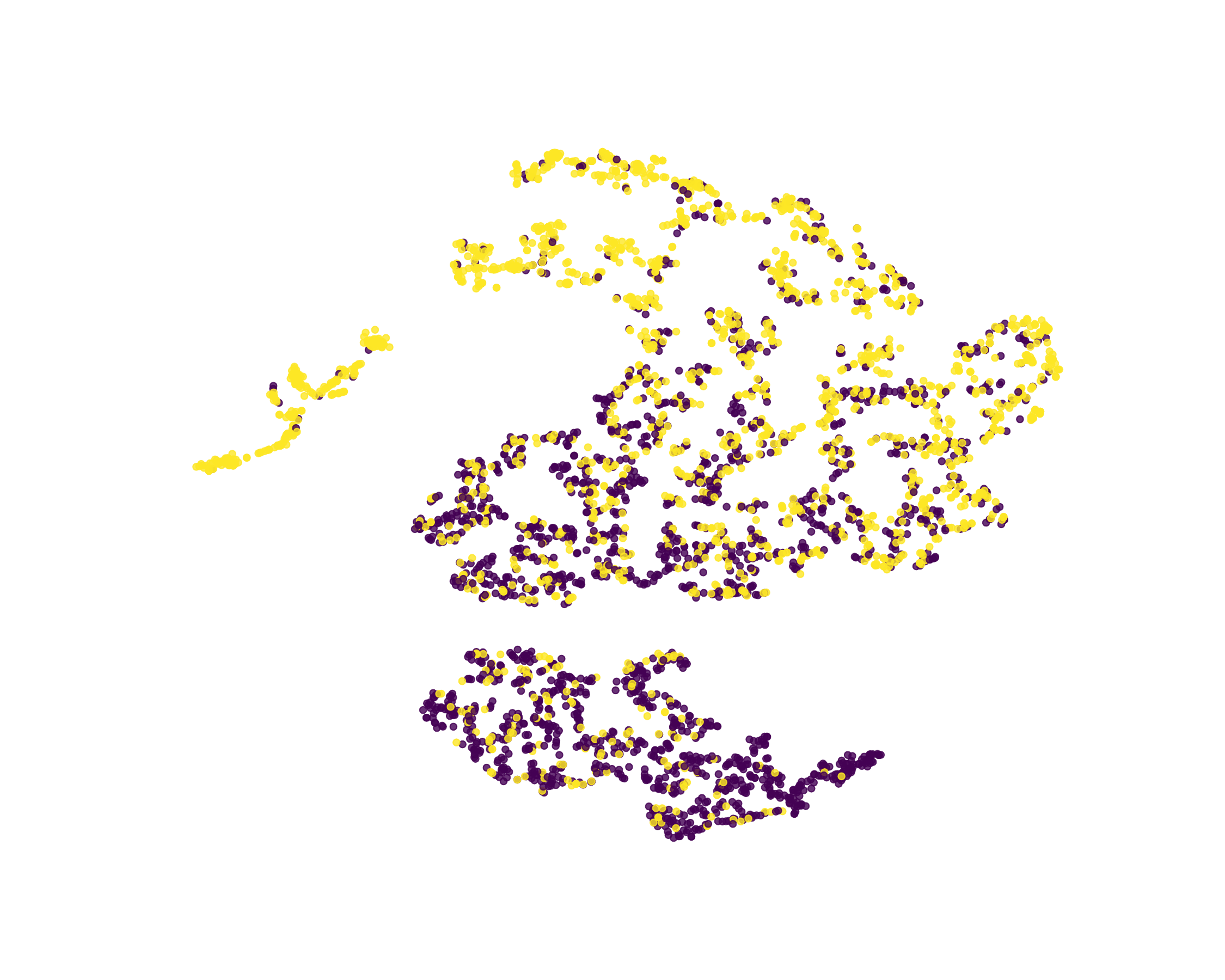}
   }
  \caption{Visualization of graphs' representation after training on NCI1.}
  \label{gc_vis}
\end{figure}

We apply TSNE \cite{tsne} to visualize the results of node and graph representations learned by graph neural networks and explain the reason why they work better with the GPL. The learned node-level or graph-level representation vectors are employed as input features in TSNE. As shown in Figure~\ref{nc_vis}, the nodes are clustered according to their class labels after being trained on the node classification task. Compared with GCN (Figure~\ref{nc_vis} (a)),  nodes' representation produced by GCN+GPL (Figure~\ref{nc_vis} (b)) is more separated by classes and densely clustered within classes. For the graph classification task, as shown in Figure~\ref{gc_vis}, after training with GPL, the learned graph features learned by SAGPool \cite{sagpool2019} are more in line with their class labels. It indicates that the proposed method facilitates the graph neural networks to produce better node-level and graph-level representation vectors. 

\begin{figure*}[!h]
  \centering
  \subfigure[]{
    \includegraphics[scale=0.25]{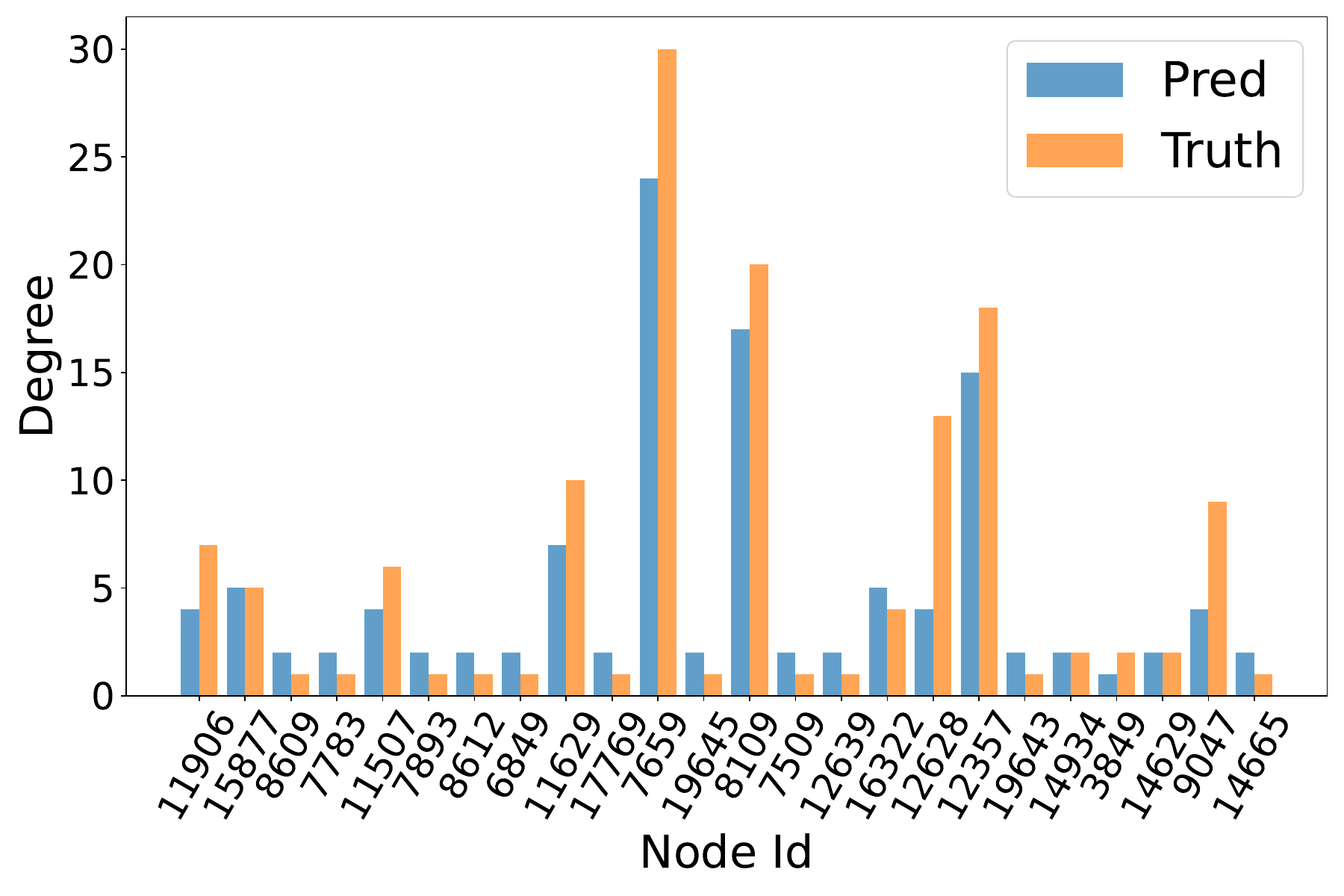}
  }
  \subfigure[]{
    \includegraphics[scale=0.25]{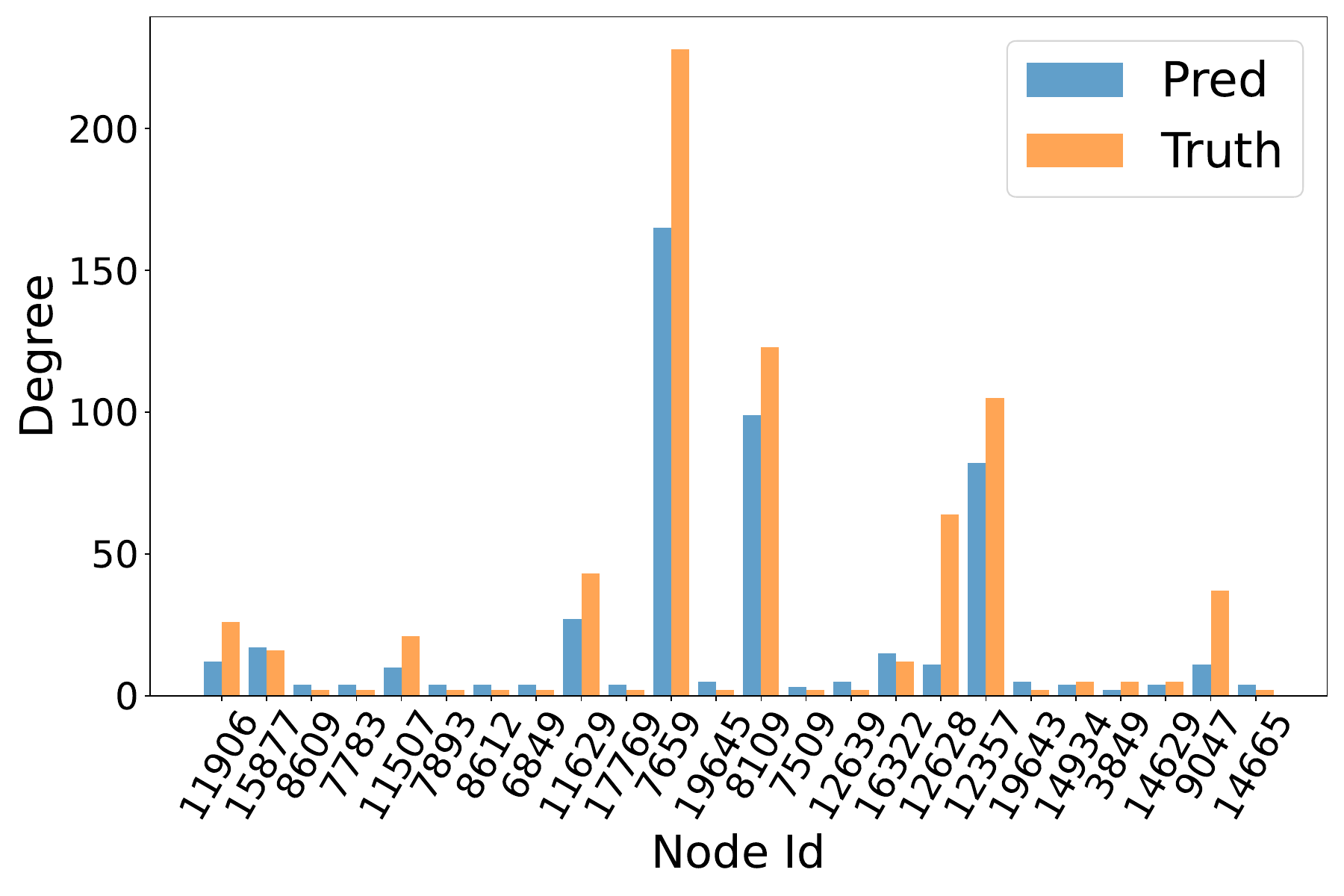}
  }
  \caption{Using node features learned by GCN+GPL on the Pubmed dataset\cite{sen2008} to predict (a) nodes' degree and (b) sum degree of nodes' neighbors.}
  \label{moti_gpl}
\end{figure*}

\begin{figure}
    \centering
    \subfigure[GCN, central node is 943.]{
        \includegraphics[scale=0.2]{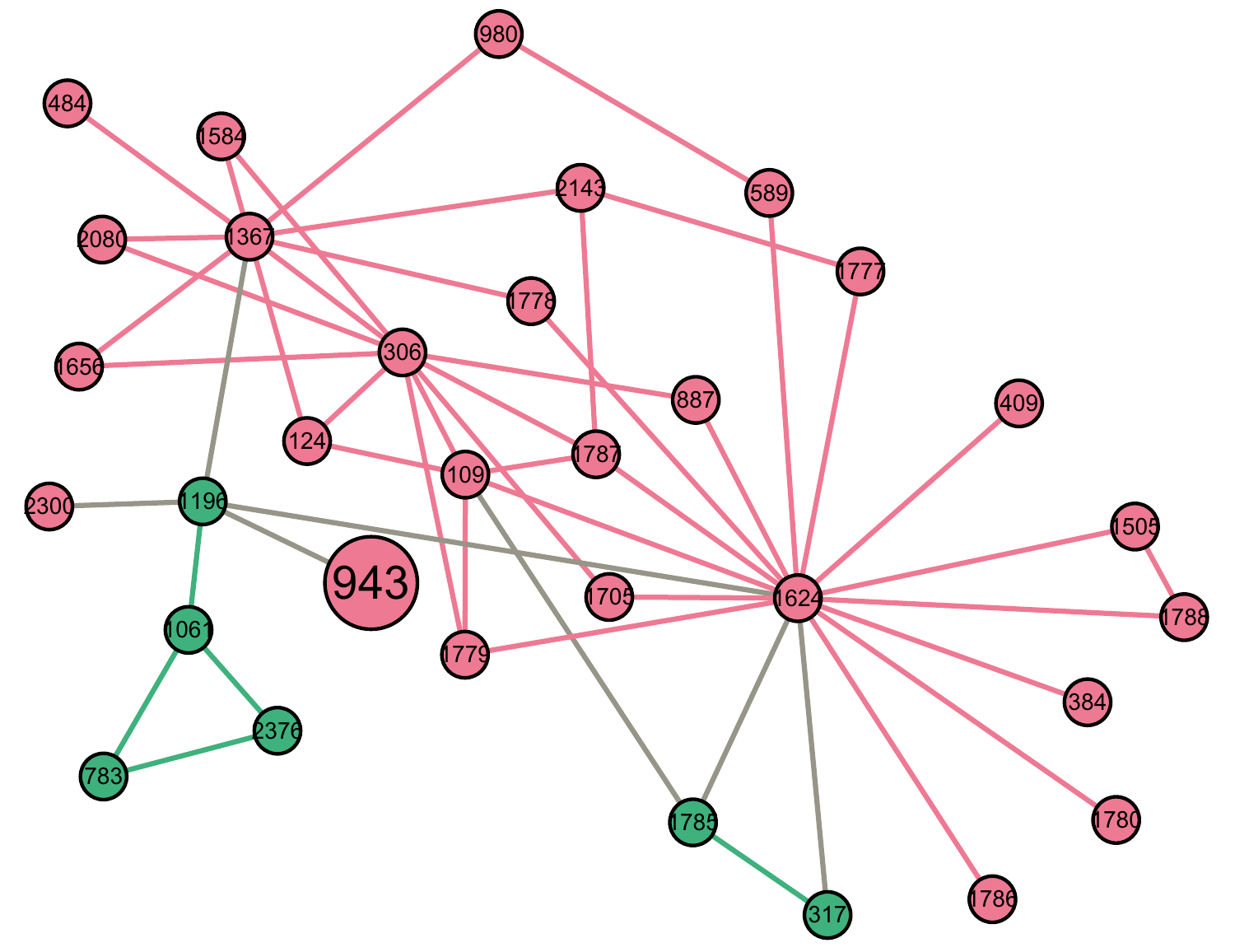}
    }
    \subfigure[GAT, central node is 578.]{
        \includegraphics[scale=0.2]{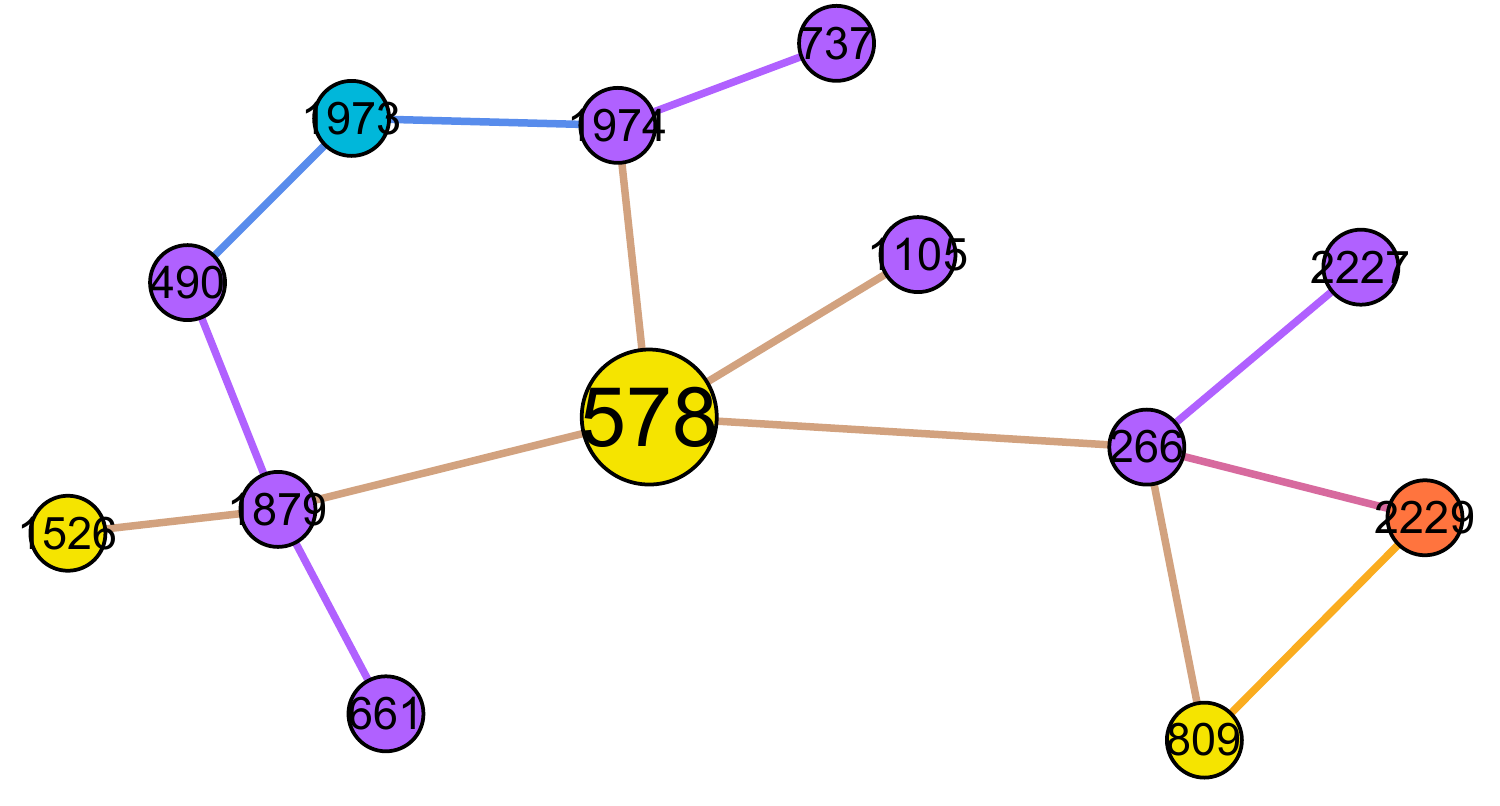}
    }
    \subfigure[GAT, central node is 787.]{
        \includegraphics[scale=0.2]{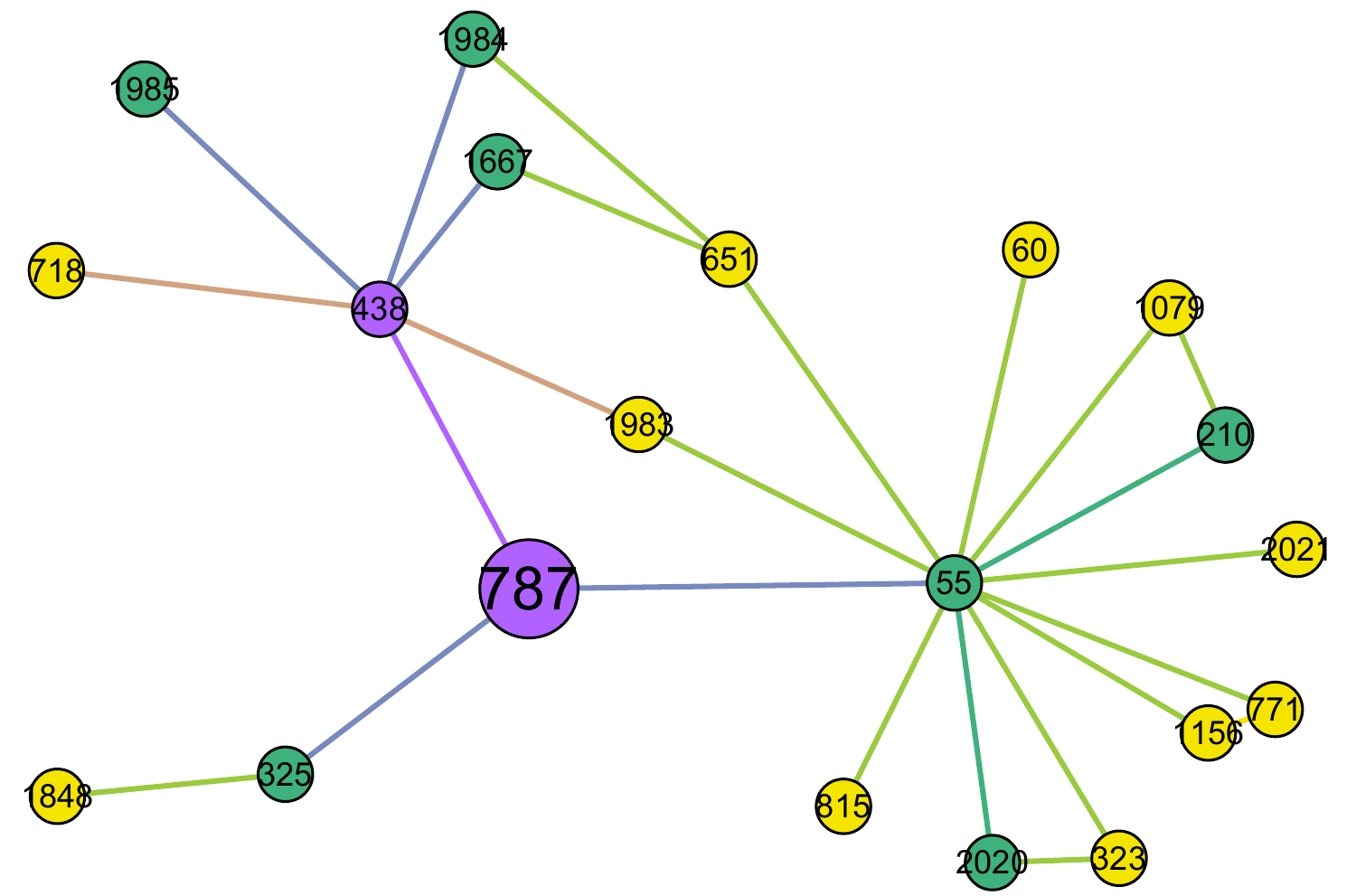}
    }
    \subfigure[GCN+GPL, central node is 943.]{
        \includegraphics[scale=0.2]{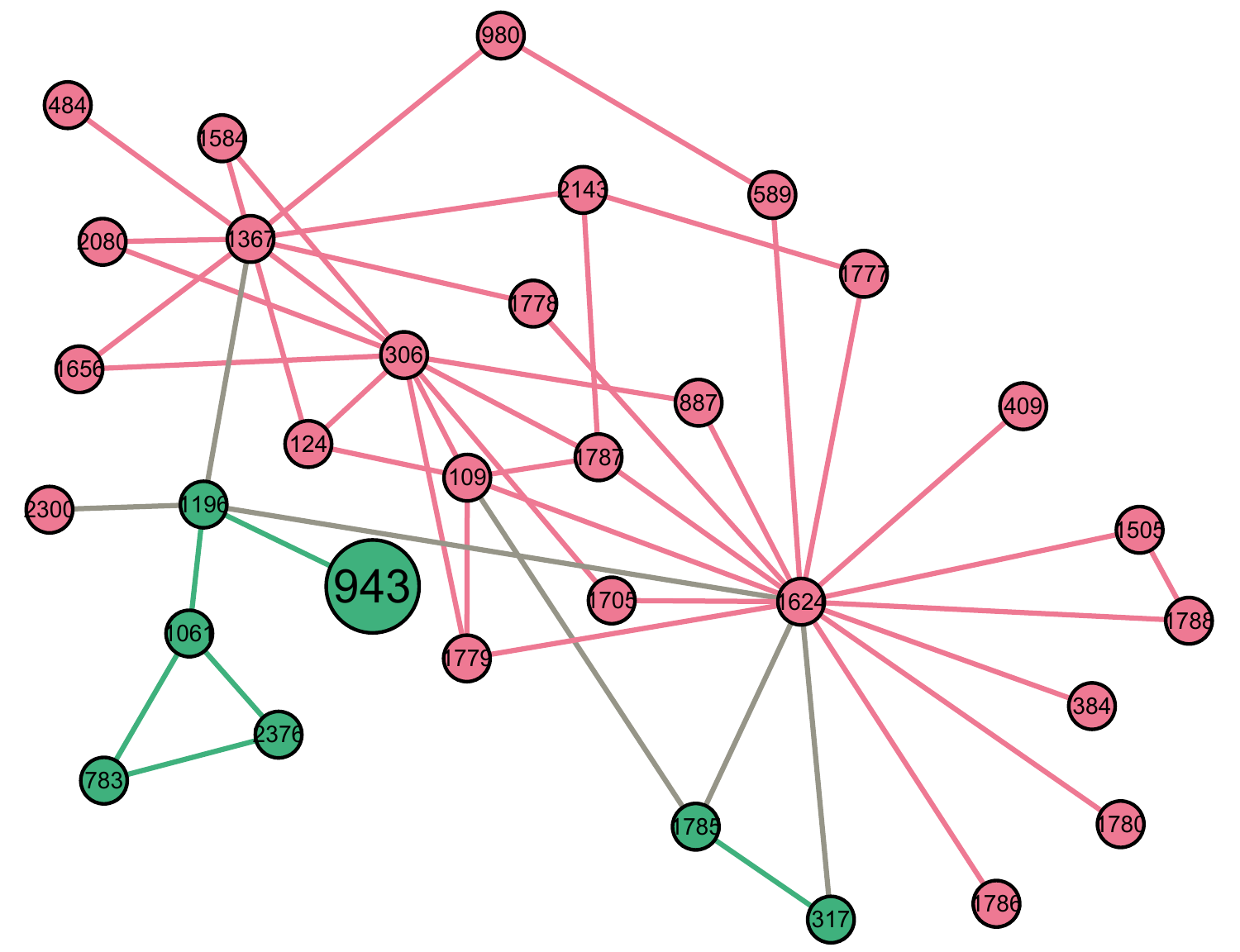}
    }
    \subfigure[GAT+GPL, central node is 578.]{
        \includegraphics[scale=0.2]{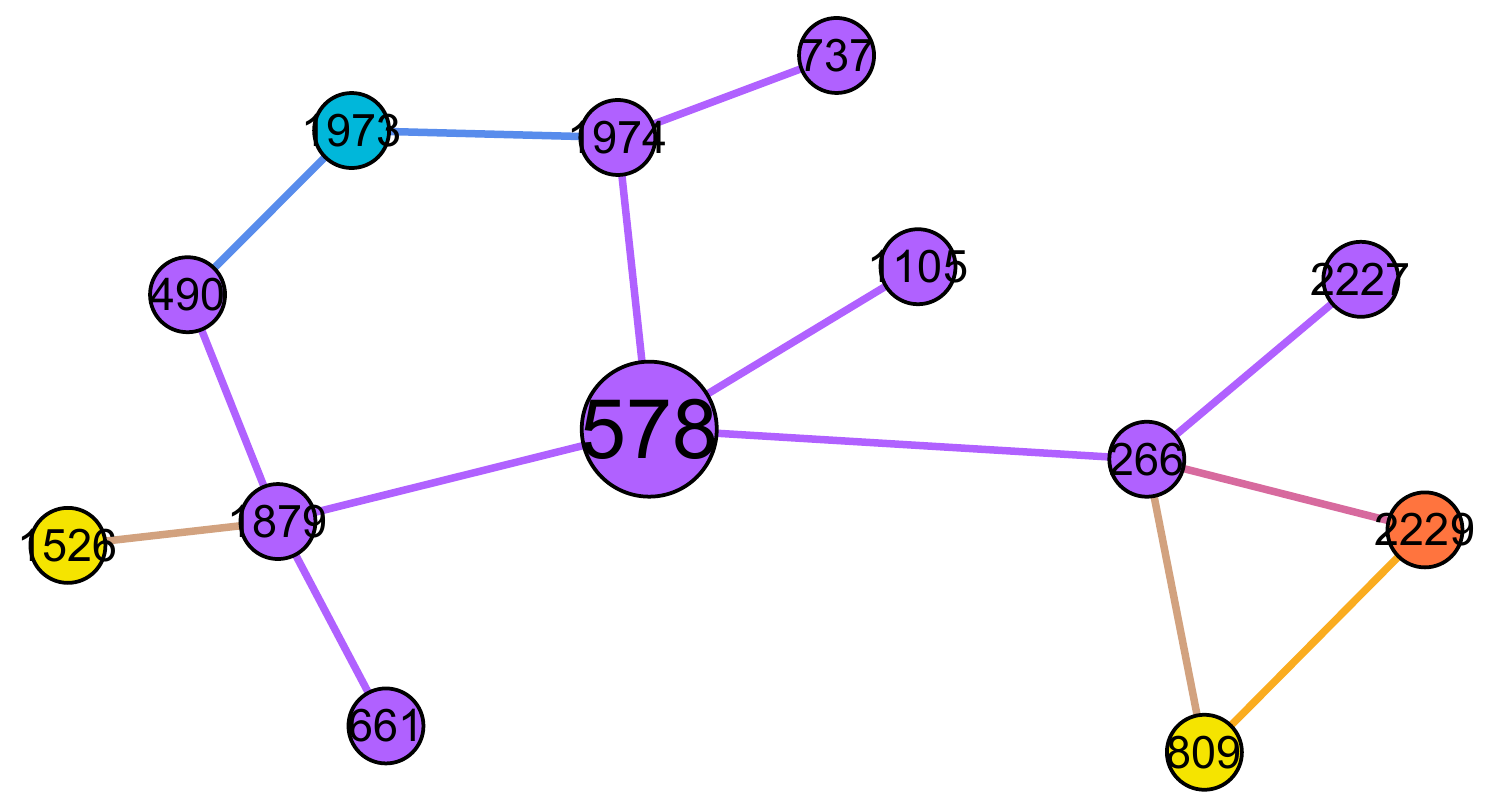}
    }
    \subfigure[GAT+GPL, central node is 787.]{
        \includegraphics[scale=0.2]{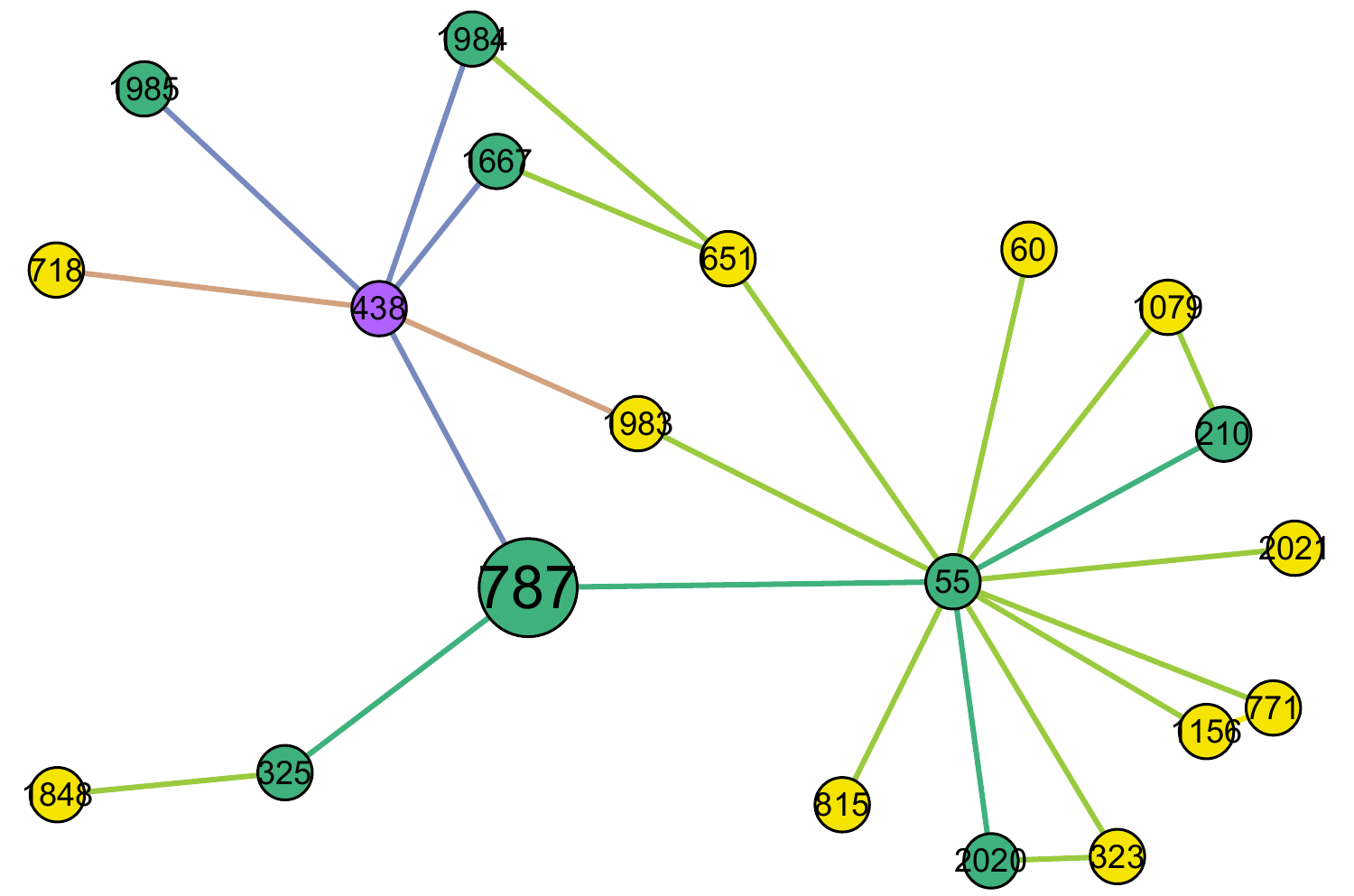}
    }
    \caption{Cases on Cora. Nodes with the same color are in the same class.}
    \label{phe}
\end{figure}

As demonstrated in Figure~\ref{motivation} (in Section Introduction),  the features learned by the GCN cannot predict nodes' degrees and the sum of their neighbors' degrees well. We fixed the learned features by GCN+GPL and trained a regression model to perform the tasks by the same experimental setups. The results are shown in Figure~\ref{moti_gpl}. The MSE and MAPE of predicting nodes' degree by GCN are 54.17 and 1.02, respectively. The MSE and MAPE by GCN+GPL are 8.12 and 0.61, reduced by 85.01\% and 40.20\%, respectively. MSE and MAPE of predicting the sum degree of nodes' neighbors by GCN+GPL are reduced by 87.11\% and 52.98\%, respectively. It can be inferred that better graph structure features have been learned in the node representations by the training method GPL.

\subsection{Case Studies}
In this subsection, we discuss the possible reasons for the improved performance brought by GPL for node classification in case studies. We compare the results by GCN and GCN+GPL, GAT and GAT+GPL on the dataset Cora. We visualize the neighboring nodes and their labels around a central node as shown in Figure~\ref{phe}. In Figure~\ref{phe}, the colors represent the categories of nodes (papers), and the bigger nodes are the central nodes. Pink: genetic algorithm; green: reinforcement learning; orange: probabilistic approach; blue: theory; purple: rule learning; yellow: case based. From Figure~\ref{phe}, we can see that the relationships between node types, graph structures, and neighboring nodes are non-trivial. In Figure~\ref{phe} (a) and (d), the ground truth of node 943 is green. The prediction result of node 943 by GCN is pink, which is affected by the two-hop hub neighbors (nodes 306 and 1624). The GCN relies on information from neighboring nodes, and the predictions of GCN suffer from the over-smooth problem. In Figure~\ref{phe}(b), the GAT predicts node 578 as class yellow might be because of the influences of two-hop yellow neighbor nodes. However, the GAT+GPL makes the right prediction.  In Figure~\ref{phe}(c), the prediction of GAT for node 787 is impacted by a one-hop purple neighborhood and predicts the center node as purple by fault. In contrast, GPL-based GCN and GAT learn more global and local graph structural information by graph structure prompts. This allows them to make correct predictions that are less dependent on local neighboring nodes and more informed by the overall structure of the graph and the nodes themselves. These cases highlight the importance of incorporating structural information of nodes and graphs into GNNs and provide evidence for the effectiveness of GPL.

\subsection{Discussions}
Most current works in GNNs focus on designing new strategies of aggregation or pooling functions \cite{gnnsurvey2022}, while still adhering to the messaging passing framework \cite{mpnn2017, pyg2019}. Some researchers have proposed specifically designed GNNs, But rare work analyzes why the performance of the current GNNs meets a bottleneck. Recent work \cite{yang2023pmlp} argues some of GNNs are not as powerful as MLPs. In contrast, this paper identifies the reason for the limitation of GNNs and focuses on improving the potential performance of existing GNNs. GPL is a generalized method that works on any graph neural network and facilitates GNNs to tackle the weakness of node or graph representation learning. GPL fulfills the theory of prompt mechanism work on graph training. Unlike graph pretraining methodologies \cite{hu2019strategies, gcc2020, gppt, graphprompt} that require carefully designed techniques and extra large datasets, GPL demonstrates that we can significantly improve the potential of current graph neural networks without additional data or high-complexity computations. In practical implications, GPL has the potential to impact the training and applications of future GNNs.

\section{Conclusion}

The current graph neural networks fail to learn a good representation of the graph structure. One reason is that the original loss is strictly related to the downstream tasks while ignoring the inherent graph structure learning. To address the issue, we propose downstream independent losses and the graph structure prompt learning (GPL) method. The performance of GPL is verified extensively on more than a dozen classic and advanced graph neural networks. The GPL significantly improves the performance for node classification, graph classification, and edge prediction tasks on over eleven representative datasets. The work provides a new perspective on the design and research of GNNs. 
The limitation of GPL in this paper is that we cannot ensure GPL constantly improves any graph neural networks from a theoretical perspective view. However, practically, it improves the performance of all the graph neural networks we have ever tried. In future work, we will explore the mathematical foundations under the GPL and explain why it leads to significant improvements. 
\section*{Acknowledgments}
This document is funded by the National Science Foundation of China (No.71971002). We also thank the support of PyG \cite{pyg2019} for our experiments.

\section*{Declaration of competing interest}
The authors declare they have no known competing financial interests or personal relationships that may influence the work.

\section*{Data availability}
All the data and codes are in https://github.com/PreckLi/graph\_prompt\_learning.

\section*{Ethical and informed consent for data used}
This paper does not contain any studies with human or animal subjects. All the human participants in the paper are informed consent. All the source datasets are publicly accessed and anonymous, and there is no offense or conflict to any human beings.

\bibliographystyle{unsrt}  
\bibliography{references}

\begin{thebibliography}{10}

\bibitem{gnnreview1}
Sergi Abadal, Akshay Jain, Robert Guirado, Jorge L{\'o}pez-Alonso, and Eduard Alarc{\'o}n.
\newblock Computing graph neural networks: A survey from algorithms to accelerators.
\newblock {\em ACM Computing Surveys (CSUR)}, 54(9):1--38, 2021.

\bibitem{knowledge2021}
Yu~Zhao, Han Zhou, Ruobing Xie, Fuzhen Zhuang, Qing Li, and Ji~Liu.
\newblock Incorporating global information in local attention for knowledge representation learning.
\newblock In {\em Findings of the Association for Computational Linguistics: ACL-IJCNLP 2021}, pages 1341--1351, 2021.

\bibitem{2022exgnn}
David Jaime~Tena Cucala, Bernardo~Cuenca Grau, Egor~V Kostylev, and Boris Motik.
\newblock Explainable gnn-based models over knowledge graphs.
\newblock In {\em International Conference on Learning Representations}, 2022.

\bibitem{textc2019}
Liang Yao, Chengsheng Mao, and Yuan Luo.
\newblock Graph convolutional networks for text classification.
\newblock In {\em Proceedings of the AAAI conference on artificial intelligence}, volume~33, pages 7370--7377, 2019.

\bibitem{2020heterogeneous4doc}
Danqing Wang, Pengfei Liu, Yining Zheng, Xipeng Qiu, and Xuanjing Huang.
\newblock Heterogeneous graph neural networks for extractive document summarization.
\newblock In {\em Proceedings of the 58th Annual Meeting of the Association for Computational Linguistics}, July 2020.

\bibitem{traffic2021}
Mengzhang Li and Zhanxing Zhu.
\newblock Spatial-temporal fusion graph neural networks for traffic flow forecasting.
\newblock In {\em Proceedings of the AAAI conference on artificial intelligence}, volume~35, pages 4189--4196, 2021.

\bibitem{trafficprediction2022}
Xu~Zhou, Yong Zhang, Zhao Li, Xing Wang, Juan Zhao, and Zhao Zhang.
\newblock Large-scale cellular traffic prediction based on graph convolutional networks with transfer learning.
\newblock {\em Neural Computing and Applications}, pages 1--11, 2022.

\bibitem{molecular2018}
Jiaxuan You, Bowen Liu, Zhitao Ying, Vijay Pande, and Jure Leskovec.
\newblock Graph convolutional policy network for goal-directed molecular graph generation.
\newblock In {\em Proceedings of Advances in neural information processing systems}, volume~31, 2018.

\bibitem{2023gambgnn}
Shoujia Zhang, Weidong Xie, Wei Li, Linjie Wang, and Chaolu Feng.
\newblock Gamb-gnn: Graph neural networks learning from gene structure relations and markov blanket ranking for cancer classification in microarray data.
\newblock {\em Chemometrics and Intelligent Laboratory Systems}, 232:104713, 2023.

\bibitem{2023ipm_gnn4recommand}
Junjie Huang, Ruobing Xie, Qi~Cao, Huawei Shen, Shaoliang Zhang, Feng Xia, and Xueqi Cheng.
\newblock Negative can be positive: Signed graph neural networks for recommendation.
\newblock {\em Information Processing \& Management}, 60(4):103403, 2023.

\bibitem{2022sent_analy}
Bin Liang, Hang Su, Lin Gui, Erik Cambria, and Ruifeng Xu.
\newblock Aspect-based sentiment analysis via affective knowledge enhanced graph convolutional networks.
\newblock {\em Knowledge-Based Systems}, 235:107643, 2022.

\bibitem{sentiment2023}
Gabriel Gomes, Ulisses Corr{\^e}a, and Larissa Freitas.
\newblock Gabsa-pt: Graph neural networks for aspect-level sentiment analysis in portuguese language.
\newblock In {\em The International FLAIRS Conference Proceedings}, volume~36, 2023.

\bibitem{gnn2016}
Micha{\"e}l Defferrard, Xavier Bresson, and Pierre Vandergheynst.
\newblock Convolutional neural networks on graphs with fast localized spectral filtering.
\newblock In {\em Proceedings of Advances in Neural Information Processing Systems}, pages 3844--3852, 2016.

\bibitem{gcn2017}
Thomas~N Kipf and Max Welling.
\newblock Semi-supervised classification with graph convolutional networks.
\newblock {\em Proceedings of International Conference on Learning Representations}, 2017.

\bibitem{sage2017}
Will Hamilton, Zhitao Ying, and Jure Leskovec.
\newblock Inductive representation learning on large graphs.
\newblock {\em Proceedings of advances in neural information processing systems}, 30, 2017.

\bibitem{gat2018}
Petar Velickovic, Guillem Cucurull, Arantxa Casanova, Adriana Romero, Pietro Lio, and Yoshua Bengio.
\newblock Graph attention networks.
\newblock {\em Proceedings of International Conference on Learning Representations}, 2018.

\bibitem{lgconv2020}
Xiangnan He, Kuan Deng, Xiang Wang, Yan Li, Yongdong Zhang, and Meng Wang.
\newblock Lightgcn: Simplifying and powering graph convolution network for recommendation.
\newblock In {\em Proceedings of the 43rd International ACM SIGIR conference on research and development in Information Retrieval}, pages 639--648, 2020.

\bibitem{transformconv2021}
Yunsheng Shi, Zhengjie Huang, Shikun Feng, Hui Zhong, Wenjin Wang, and Yu~Sun.
\newblock Masked label prediction: Unified message passing model for semi-supervised classification.
\newblock {\em Proceedings of International Joint Conference on Artificial Intelligence}, 2021.

\bibitem{ARMA2021}
Filippo~Maria Bianchi, Daniele Grattarola, Lorenzo Livi, and Cesare Alippi.
\newblock Graph neural networks with convolutional arma filters.
\newblock {\em IEEE Transactions on Pattern Analysis and Machine Intelligence}, 2021.

\bibitem{2022fusedgat}
Hengrui Zhang, Zhongming Yu, Guohao Dai, Guyue Huang, Yufei Ding, Yuan Xie, and Yu~Wang.
\newblock Understanding gnn computational graph: A coordinated computation, io, and memory perspective.
\newblock In {\em Proceedings of Machine Learning and Systems}, volume~4, pages 467--484, 2022.

\bibitem{2022anti-symmetric}
Alessio Gravina, Davide Bacciu, and Claudio Gallicchio.
\newblock Anti-symmetric {DGN}: a stable architecture for deep graph networks.
\newblock In {\em The Eleventh International Conference on Learning Representations}, 2023.

\bibitem{gin2018}
Keyulu Xu, Weihua Hu, Jure Leskovec, and Stefanie Jegelka.
\newblock How powerful are graph neural networks?
\newblock In {\em Proceedings of International Conference on Learning Representations}, 2018.

\bibitem{sortpool2018}
Muhan Zhang, Zhicheng Cui, Marion Neumann, and Yixin Chen.
\newblock An end-to-end deep learning architecture for graph classification.
\newblock In {\em Proceedings of the AAAI conference on artificial intelligence}, volume~32, 2018.

\bibitem{diffpool2019}
Zhitao Ying, Jiaxuan You, Christopher Morris, Xiang Ren, Will Hamilton, and Jure Leskovec.
\newblock Hierarchical graph representation learning with differentiable pooling.
\newblock In {\em Proceedings of Advances in neural information processing systems}, volume~31, 2018.

\bibitem{topkpool2019}
Boris Knyazev, Graham~W Taylor, and Mohamed Amer.
\newblock Understanding attention and generalization in graph neural networks.
\newblock {\em Proceedings of Advances in neural information processing systems}, 32, 2019.

\bibitem{sagpool2019}
Junhyun Lee, Inyeop Lee, and Jaewoo Kang.
\newblock Self-attention graph pooling.
\newblock In {\em Proceedings of international conference on machine learning}, pages 3734--3743. PMLR, 2019.

\bibitem{edgepool2019}
Frederik Diehl, Thomas Brunner, Michael~Truong Le, and Alois Knoll.
\newblock Towards graph pooling by edge contraction.
\newblock In {\em ICML 2019 workshop on learning and reasoning with graph-structured data}, 2019.

\bibitem{asapool2020}
Ekagra Ranjan, Soumya Sanyal, and Partha Talukdar.
\newblock Asap: Adaptive structure aware pooling for learning hierarchical graph representations.
\newblock In {\em Proceedings of the AAAI Conference on Artificial Intelligence}, volume~34, pages 5470--5477, 2020.

\bibitem{MEWISPool2021}
Amirhossein Nouranizadeh, Mohammadjavad Matinkia, Mohammad Rahmati, and Reza Safabakhsh.
\newblock Maximum entropy weighted independent set pooling for graph neural networks.
\newblock In {\em Proceedings of Advances in neural information processing systems}, 2021.

\bibitem{GPS2022}
Ladislav Ramp{\'a}{\v{s}}ek, Michael Galkin, Vijay~Prakash Dwivedi, Anh~Tuan Luu, Guy Wolf, and Dominique Beaini.
\newblock Recipe for a general, powerful, scalable graph transformer.
\newblock In {\em Advances in Neural Information Processing Systems}, volume~35, pages 14501--14515, 2022.

\bibitem{hu2019strategies}
Weihua Hu, Bowen Liu, Joseph Gomes, Marinka Zitnik, Percy Liang, Vijay Pande, and Jure Leskovec.
\newblock Strategies for pre-training graph neural networks.
\newblock In {\em Proceedings of International Conference on Learning Representations}, 2020.

\bibitem{gcc2020}
Jiezhong Qiu, Qibin Chen, Yuxiao Dong, Jing Zhang, Hongxia Yang, Ming Ding, Kuansan Wang, and Jie Tang.
\newblock Gcc: Graph contrastive coding for graph neural network pre-training.
\newblock In {\em Proceedings of the 26th ACM SIGKDD International Conference on Knowledge Discovery \& Data Mining}, pages 1150--1160, 2020.

\bibitem{gptgnn2020}
Ziniu Hu, Yuxiao Dong, Kuansan Wang, Kai-Wei Chang, and Yizhou Sun.
\newblock Gpt-gnn: Generative pre-training of graph neural networks.
\newblock In {\em Proceedings of the 26th ACM SIGKDD International Conference on Knowledge Discovery \& Data Mining}, pages 1857--1867, 2020.

\bibitem{L2P2021}
Yuanfu Lu, Xunqiang Jiang, Yuan Fang, and Chuan Shi.
\newblock Learning to pre-train graph neural networks.
\newblock In {\em Proceedings of the AAAI Conference on Artificial Intelligence}, volume~35, pages 4276--4284, 2021.

\bibitem{sugar2021}
Qingyun Sun, Jianxin Li, Hao Peng, Jia Wu, Yuanxing Ning, Philip~S Yu, and Lifang He.
\newblock Sugar: Subgraph neural network with reinforcement pooling and self-supervised mutual information mechanism.
\newblock In {\em Proceedings of the Web Conference 2021}, pages 2081--2091, 2021.

\bibitem{metagnn}
Aravind Sankar, Xinyang Zhang, and Kevin Chen-Chuan Chang.
\newblock Meta-gnn: Metagraph neural network for semi-supervised learning in attributed heterogeneous information networks.
\newblock In {\em Proceedings of the 2019 IEEE/ACM International Conference on Advances in Social Networks Analysis and Mining}, pages 137--144, 2019.

\bibitem{yang2023pmlp}
Chenxiao Yang, Qitian Wu, Jiahua Wang, and Junchi Yan.
\newblock Graph neural networks are inherently good generalizers: Insights by bridging gnns and mlps.
\newblock In {\em International Conference on Learning Representations}, 2023.

\bibitem{gnnreview2}
Zonghan Wu, Shirui Pan, Fengwen Chen, Guodong Long, Chengqi Zhang, and S~Yu Philip.
\newblock A comprehensive survey on graph neural networks.
\newblock {\em IEEE transactions on neural networks and learning systems}, 32(1):4--24, 2020.

\bibitem{gnnsurvey2022}
Yu~Zhou, Haixia Zheng, Xin Huang, Shufeng Hao, Dengao Li, and Jumin Zhao.
\newblock Graph neural networks: Taxonomy, advances, and trends.
\newblock {\em ACM Transactions on Intelligent Systems and Technology (TIST)}, 13(1):1--54, 2022.

\bibitem{gnnreview3}
Jie Zhou, Ganqu Cui, Shengding Hu, Zhengyan Zhang, Cheng Yang, Zhiyuan Liu, Lifeng Wang, Changcheng Li, and Maosong Sun.
\newblock Graph neural networks: A review of methods and applications.
\newblock {\em AI Open}, 1:57--81, 2020.

\bibitem{sen2008}
Prithviraj Sen, Galileo Namata, Mustafa Bilgic, Lise Getoor, Brian Galligher, and Tina Eliassi-Rad.
\newblock Collective classification in network data.
\newblock {\em AI magazine}, 29(3):93--93, 2008.

\bibitem{schick2020coling}
Timo Schick, Helmut Schmid, and Hinrich Sch{\"u}tze.
\newblock Automatically identifying words that can serve as labels for few-shot text classification.
\newblock In {\em Proceedings of the 28th International Conference on Computational Linguistics}, pages 5569--5578, 2020.

\bibitem{2021graphprompt}
Jiayou Zhang, Zhirui Wang, Shizhuo Zhang, Megh~Manoj Bhalerao, Yucong Liu, Dawei Zhu, and Sheng Wang.
\newblock Graphprompt: Biomedical entity normalization using graph-based prompt templates.
\newblock In {\em Proceedings of the ACM Web Conference 2023}, 2021.

\bibitem{promptreview2021}
Pengfei Liu, Weizhe Yuan, Jinlan Fu, Zhengbao Jiang, Hiroaki Hayashi, and Graham Neubig.
\newblock Pre-train, prompt, and predict: A systematic survey of prompting methods in natural language processing.
\newblock {\em ACM Computing Surveys}, 55(9):1--35, 2023.

\bibitem{softprompt2021}
Xiao Liu, Kaixuan Ji, Yicheng Fu, Weng Tam, Zhengxiao Du, Zhilin Yang, and Jie Tang.
\newblock P-tuning: Prompt tuning can be comparable to fine-tuning across scales and tasks.
\newblock In {\em Proceedings of the 60th Annual Meeting of the Association for Computational Linguistics}, pages 61--68, 2022.

\bibitem{gppt}
Mingchen Sun, Kaixiong Zhou, Xin He, Ying Wang, and Xin Wang.
\newblock Gppt: Graph pre-training and prompt tuning to generalize graph neural networks.
\newblock In {\em Proceedings of the 28th ACM SIGKDD Conference on Knowledge Discovery and Data Mining}, pages 1717--1727, 2022.

\bibitem{graphprompt}
Zemin Liu, Xingtong Yu, Yuan Fang, and Xinming Zhang.
\newblock Graphprompt: Unifying pre-training and downstream tasks for graph neural networks.
\newblock In {\em Proceedings of the ACM Web Conference 2023}, 2023.

\bibitem{2022location_aware}
Zhaohui Wang, Qi~Cao, Huawei Shen, Bingbing Xu, Keting Cen, and Xueqi Cheng.
\newblock Location-aware convolutional neural networks for graph classification.
\newblock {\em Neural Networks}, 155:74--83, 2022.

\bibitem{bi2023mm}
Wendong Bi, Lun Du, Qiang Fu, Yanlin Wang, Shi Han, and Dongmei Zhang.
\newblock Mm-gnn: Mix-moment graph neural network towards modeling neighborhood feature distribution.
\newblock In {\em Proceedings of the Sixteenth ACM International Conference on Web Search and Data Mining}, pages 132--140, 2023.

\bibitem{zhang2023iea}
Peiliang Zhang, Jiatao Chen, Chao Che, Liang Zhang, Bo~Jin, and Yongjun Zhu.
\newblock Iea-gnn: Anchor-aware graph neural network fused with information entropy for node classification and link prediction.
\newblock {\em Information Sciences}, 634:665--676, 2023.

\bibitem{deepgcn2019}
Guohao Li, Matthias Muller, Ali Thabet, and Bernard Ghanem.
\newblock Deepgcns: Can gcns go as deep as cnns?
\newblock In {\em Proceedings of the IEEE/CVF international conference on computer vision}, pages 9267--9276, 2019.

\bibitem{ppnp2019}
Bingbing Xu, Huawei Shen, Qi~Cao, Keting Cen, and Xueqi Cheng.
\newblock Graph convolutional networks using heat kernel for semi-supervised learning.
\newblock In {\em Proceedings of International Joint Conference on Artificial Intelligence}, 2019.

\bibitem{2023casangcl}
Zixi Zheng, Yanyan Tan, Hong Wang, Shengpeng Yu, Tianyu Liu, and Cheng Liang.
\newblock Casangcl: pre-training and fine-tuning model based on cascaded attention network and graph contrastive learning for molecular property prediction.
\newblock {\em Briefings in Bioinformatics}, 24(1):bbac566, 2023.

\bibitem{2023molebert}
Jun Xia, Chengshuai Zhao, Bozhen Hu, Zhangyang Gao, Cheng Tan, Yue Liu, Siyuan Li, and Stan~Z Li.
\newblock Mole-bert: Rethinking pre-training graph neural networks for molecules.
\newblock In {\em Proceedings of International Conference on Learning Representations}, 2023.

\bibitem{bert2019}
Jacob Devlin Ming-Wei~Chang Kenton and Lee~Kristina Toutanova.
\newblock Bert: Pre-training of deep bidirectional transformers for language understanding.
\newblock In {\em Proceedings of NAACL-HLT}, pages 4171--4186, 2019.

\bibitem{mpnn2017}
Justin Gilmer, Samuel~S Schoenholz, Patrick~F Riley, Oriol Vinyals, and George~E Dahl.
\newblock Neural message passing for quantum chemistry.
\newblock In {\em International conference on machine learning}, pages 1263--1272. PMLR, 2017.

\bibitem{2018pitfall}
Oleksandr Shchur, Maximilian Mumme, Aleksandar Bojchevski, and Stephan G{\"u}nnemann.
\newblock Pitfalls of graph neural network evaluation.
\newblock In {\em Relational Representation Learning Workshop, NeurIPS 2018}, 2018.

\bibitem{dblp2017}
Aleksandar Bojchevski and Stephan G{\"u}nnemann.
\newblock Deep gaussian embedding of graphs: Unsupervised inductive learning via ranking.
\newblock In {\em Proceedings of International Conference on Learning Representations}, 2018.

\bibitem{1991mutag}
Asim~Kumar Debnath, Rosa~L Lopez~de Compadre, Gargi Debnath, Alan~J Shusterman, and Corwin Hansch.
\newblock Structure-activity relationship of mutagenic aromatic and heteroaromatic nitro compounds. correlation with molecular orbital energies and hydrophobicity.
\newblock {\em Journal of medicinal chemistry}, 34(2):786--797, 1991.

\bibitem{2005proteins}
Karsten~M Borgwardt, Cheng~Soon Ong, Stefan Sch{\"o}nauer, SVN Vishwanathan, Alex~J Smola, and Hans-Peter Kriegel.
\newblock Protein function prediction via graph kernels.
\newblock {\em Bioinformatics}, 21(suppl\_1):i47--i56, 2005.

\bibitem{dd2003}
Paul~D Dobson and Andrew~J Doig.
\newblock Distinguishing enzyme structures from non-enzymes without alignments.
\newblock {\em Journal of molecular biology}, 330(4):771--783, 2003.

\bibitem{NCI2008}
Nikil Wale, Ian~A Watson, and George Karypis.
\newblock Comparison of descriptor spaces for chemical compound retrieval and classification.
\newblock {\em Knowledge and Information Systems}, 14(3):347--375, 2008.

\bibitem{transformer2017}
Ashish Vaswani, Noam Shazeer, Niki Parmar, Jakob Uszkoreit, Llion Jones, Aidan~N Gomez, \L~ukasz Kaiser, and Illia Polosukhin.
\newblock Attention is all you need.
\newblock In {\em Proceedings of the 31st Advances in Neural Information Processing Systems}, pages 5998--6008, 2017.

\bibitem{pyg2019}
Matthias Fey and Jan~E. Lenssen.
\newblock Fast graph representation learning with {PyTorch Geometric}.
\newblock In {\em Proceedings of ICLR Workshop on Representation Learning on Graphs and Manifolds}, 2019.

\bibitem{2021dgnn}
Kezhao Huang, Jidong Zhai, Zhen Zheng, Youngmin Yi, and Xipeng Shen.
\newblock Understanding and bridging the gaps in current gnn performance optimizations.
\newblock In {\em Proceedings of the 26th ACM SIGPLAN Symposium on Principles and Practice of Parallel Programming}, pages 119--132, 2021.

\bibitem{tsne}
Laurens Van~der Maaten and Geoffrey Hinton.
\newblock Visualizing data using t-sne.
\newblock {\em Journal of Machine Learning Research}, 9(86):2579--2605, 2008.

\end{thebibliography}

\end{document}